\documentclass[]{prime_research}

\usepackage{mathpazo}   
\usepackage{minitoc}
\usepackage{url}
\usepackage{amssymb}
\usepackage[utf8]{inputenc}
\usepackage{microtype}
\usepackage{booktabs}
\usepackage{pifont} 
\usepackage{multirow}
\usepackage{makecell}
\usepackage{paralist}
\usepackage{xspace}
\usepackage{color}
\usepackage{xcolor}
\usepackage{colortbl}
\usepackage{adjustbox}
\usepackage{hyperref} 
\usepackage[edges]{forest}
\usepackage{tikz} 
\usepackage{caption}
\usepackage{amsfonts}
\usepackage[toc,page,header]{appendix}
\usepackage[utf8]{inputenc}
\usepackage{bbm}
\usepackage{enumitem}
\usepackage{mathtools}
\usepackage{wrapfig}
\usepackage{float}
\usepackage{caption}

\definecolor{seedc}{RGB}{7, 92, 173}

\newcommand{\name}[1]{PrimeSim}

\newcommand{\hardware}[1]{G2}

\renewcommand{\paragraph}[1]{\vspace{0.1em}\noindent\textbf{#1}}

\title{PrimeSim: A Generalist World Model Simulator from Large-Scale Cross-Embodiment Demonstrations}

\author[1]{Travis Davies}
\author[1,2]{Wentao Ma}
\author[1,3]{Lihao Guo}
\author[1,4]{Tengfei Ma}
\author[1,5]{Xiaowen Chang}
\author[7]{Jiafeng Xu}
\author[1,3,8]{Shilong Wang}
\author[1,6]{Yiqi Deng}
\author[1,7]{Shaowen He}
\author[1,7]{Xiaoqi Li}
\author[1]{Qian Zhu}
\author[1]{Ran Cheng}
\author[1,7]{Hao Dong}

\affiliation[1]{PrimeBot Research}
\affiliation[2]{Southeast University}
\affiliation[3]{Shanghai Jiao Tong University}
\affiliation[4]{The Chinese University of Hong Kong}
\affiliation[5]{University of Chinese Academy of Sciences}
\affiliation[6]{Beihang University}
\affiliation[7]{Peking University}
\affiliation[8]{Shanghai Innovation Institute}
 
\abstract{
A robot world model that serves as a simulator must do more than generate plausible videos: it must faithfully follow commanded actions, maintain consistency across camera views, and sustain coherent long-horizon rollouts.
We present \textbf{PrimeSim}, an action-conditioned, multiview world model simulator that addresses these challenges through progressive midtraining and posttraining.
Starting from a pretrained video generator, we midtrain on 30,110 hours of robot demonstrations spanning 22 platforms.
To improve \textit{action following}, we introduce a unified cross-embodiment action representation and \textbf{frame-aligned dual-path conditioning} to strengthen action--frame correspondence.
To maintain \textit{multiview consistency}, we develop \textbf{view-aware interleaved attention} with explicit camera identities, enabling cross-view information exchange while preserving view-specific representations.
We further introduce a \textbf{fast-video, slow-memory posttraining} framework that transforms the bidirectional video generator into an efficient autoregressive simulator.
By combining fast causal generation, self-rollout training, and hierarchical temporal memory, PrimeSim mitigates error accumulation from self-generated histories and supports coherent long-horizon simulation.
PrimeSim runs at 10.1 FPS, remains stable over long-horizon rollouts on rigid and deformable manipulation tasks, and transfers to held-out platforms within 100 to 200 gradient steps where the pretrained backbone does not.
Against five open-source action-conditioned world models it leads four of the five WorldSimProbe suites and posts the best overall EWMBench score, while fine-grained local calibration still trails the field; its rollouts are faithful enough to support GRPO-based policy optimization.
We release the model weights, inference code, and training code.
}

\checkdata[Project Page]{
\url{https://qyrotech.github.io/PrimeSim-website/}
}
\correspondence{\email{travisdavies@qiyuan-robot.com}}

\begin{document}
\maketitle

\section{Introduction}
\label{sec:intro}

\begin{figure*}[!t]
  \centering
  \includegraphics[width=\textwidth]{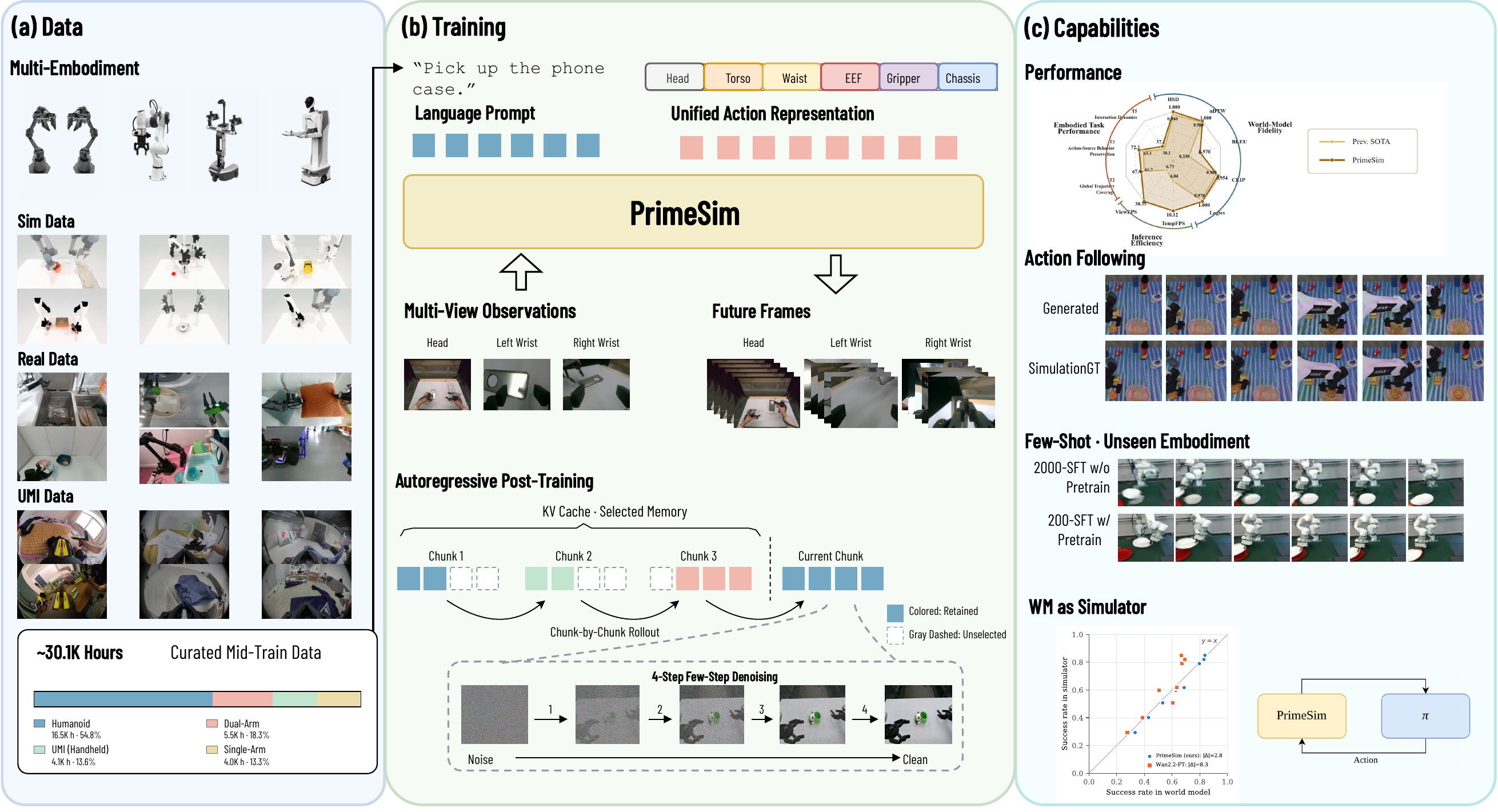}
  \caption{PrimeSim. \textbf{(a)} The midtraining corpus: roughly
  30.1K hours of curated robot data spanning humanoid, dual-arm, single-arm and
  handheld UMI~\cite{umi,hifi-umi} platforms, mixing real and simulated footage. \textbf{(b)} The
  model is conditioned on a language prompt, a unified action representation over
  head, torso, waist, end-effector, gripper and chassis, and three camera views,
  and predicts future frames; post-training makes generation autoregressive with
  a selected KV-cache memory at four denoising steps. \textbf{(c)} Capabilities:
  fidelity against prior systems, action following measured against simulation
  ground truth, few-shot adaptation to an unseen embodiment, and use as a
  simulator: success rate in PrimeSim predicts success rate in the LIBERO~\cite{libero} physics
    simulator with $r=0.99$, so policies can be trained in the model and
    the result carried back to the robot.}
  \label{fig:teaser}
\end{figure*}

A world model that renders the consequences of a policy's actions can serve as a
simulator: rollouts become parallel, repeatable, and free of the risk of damaging
hardware. This sidesteps both the danger and the sample cost of online training
on real hardware \cite{rlhf, ppo, rl_survey2024, sop}, and avoids the sim-to-real
gap \cite{soma, digital_cousin} and the per-environment engineering that
hand-built physics simulators demand. The idea runs from model-based
reinforcement learning, where agents learn from imagined rollouts alone
\cite{wm, dreamer, dreamerv3, daydreamer}, through interactive video generation
for games and open-ended worlds \cite{diamond, genie, matrix, gamengen, yume-15,
hunyuan-gamegraft, echo-joy}, to action-conditioned robot world models
\cite{ctrl-world, dreamdojo, genie-envisioner, lingbot-world2}. What such a model
must do is narrower than generating plausible video: it must adhere to the actions
a policy issues, keep several camera views consistent, and stay coherent over
rollouts far longer than the window it was trained on. We present PrimeSim, a
world model built as a robot simulator that meets all three requirements.

To enable faithful action following and consistent multiview generation, we adapt a pretrained video model to robot dynamics through \textbf{midtraining} on 30,110 hours of demonstrations spanning 22 robot platforms. For \textit{action following}, we unify heterogeneous robot controls into a shared action representation and introduce \textbf{frame-aligned dual-path action conditioning}, combining temporally aligned attention with direct action injection to establish precise correspondence between robot controls and generated motion. 
For \textit{multiview consistency}, we introduce \textbf{view-aware interleaved attention}, enabling cross-view information exchange while preserving camera-specific representations through explicit camera identities. Through joint conditional flow-matching training, these designs allow a single model to learn action-conditioned, multiview robot dynamics across diverse embodiments, establishing the foundation for an interactive world model simulator.

While midtraining establishes action-conditioned multiview generation within a fixed temporal window, turning the model into a practical robot simulator introduces a fundamental challenge: autoregressive simulation requires continuously conditioning on the model's own imperfect predictions, whereas conventional video generators are trained on fixed-length, ground-truth contexts. This mismatch leads to error accumulation over extended rollouts and is compounded by the computational cost of iterative video generation. To address these challenges, we develop a \textbf{fast-video, slow-memory posttraining} framework that progressively transforms the bidirectional video generator into an efficient, long-horizon autoregressive simulator. First, we enable \textit{fast generation} through chunk-causal adaptation and four-step consistency distillation, making sequential simulation and subsequent self-rollout training computationally tractable. Second, we improve \textit{self-conditioned stability} through self-rollout training with asymmetric distribution matching against the frozen bidirectional teacher, allowing the model to learn from its own generated histories rather than relying solely on ground-truth context. Third, we introduce \textbf{hierarchical temporal memory} to support long-horizon coherence, preserving the initial scene, sparse long-term history, and dense recent motion within a bounded KV cache. Together, these components enable efficient autoregressive simulation that remains robust to self-generated context and scales beyond the original training window.

Together, these designs enable PrimeSim to generate temporally coherent rollouts up to 105 seconds at 10.1 FPS, supporting diverse robotic manipulation tasks involving both rigid and deformable objects. Compared with five open-source action-conditioned world models, PrimeSim achieves
state-of-the-art performance on WorldSimProbe \cite{worldsimprobe}, leading four of its five suites:
global trajectory coverage by 10.0\%, action-source behavior preservation by 41.8\%, and
interaction dynamics by 48.4\%, each measured against the strongest competing model on that
axis, with interaction grounding also led at 85.7. On EWMBench, PrimeSim reaches the best overall score of 6.09, ahead of both GE-Sim variants
(5.82 and 5.79) and every instruction-conditioned model, and is best on five of the eight
dimensions, including perfect scores on high-level semantics, trajectory alignment and
logical consistency, with action semantics (BLEU) 72.7\% above the strongest GE-Sim variant. Beyond benchmark fidelity, we demonstrate the practical utility of PrimeSim as a learned simulator for GRPO-based policy optimization~\cite{grpo}, enabling policy improvement through model-generated interactions. We release the model weights, inference code, and training code to support further research.

\noindent\textbf{Our main contributions are summarized as follows:}
\begin{itemize}
    \item We present \textbf{PrimeSim}, an action-conditioned,
    multiview world model simulator that transforms a pretrained
    video generator into an interactive robot world model
    through a progressive midtraining and posttraining framework.

    \item We develop a large-scale \textbf{midtraining framework}
    using 30,110 hours of demonstrations across 22 robot platforms.
    With unified cross-embodiment action representations,
    frame-aligned dual-path conditioning, and view-aware
    interleaved attention, PrimeSim learns action-faithful
    and multiview-consistent robot dynamics across diverse
    embodiments.

    \item We introduce a \textbf{fast-video, slow-memory
    posttraining framework} that combines efficient causal
    generation, self-rollout training with asymmetric
    distribution matching, and hierarchical temporal memory.
    This enables fast, self-conditioned autoregressive
    simulation while mitigating long-horizon error accumulation
    beyond the original training window.

    \item We demonstrate \textbf{strong performance} across diverse rigid and
    deformable object manipulation tasks: 10.1 FPS at four sampling steps,
    coherent rollouts of up to 105 seconds, \textbf{6.09 overall on EWMBench}
    (first), and \textbf{first on four of five} WorldSimProbe suites against five
    open-source action-conditioned world models, retaining the lead on two of the
    three action-realisation axes and interaction dynamics. We further
    validate PrimeSim as a learned simulator for GRPO-based policy optimization.
\end{itemize} 
\section{Data Engine}
\label{sec:data_engine}

Our goal for training data was to learn from a massive and diverse pool of settings while keeping our representations simple. To this end, we assembled 30{,}110 hours of robot demonstration data from a mix of open-source and proprietary sources. Dataset statistics are shown in Fig.~\ref{fig:overview}, and an overview of the datasets used is provided in Table~\ref{tab:datasets_frames}. In general, we include any demonstration data that can be converted into our unified format with reasonable effort, whether humanoid, dual-arm, single-arm, UMI \cite{umi}, or simulated. We describe the conversion pipeline, quality filtering, and dataset composition below and in the appendix.

\paragraph{Robot Data Pooling.} Motivated by the low cost of data collection, recent state-of-the-art methods pretrain on large egocentric human videos with a smaller ratio of robot data \cite{lingbot-va, dreamdojo, cwm}. However, aligning robot actions to latent actions or predicted hand-to-end-effector poses adds additional complexity. Since pooled robot datasets, open- and closed-source together, now largely match the scale of the egocentric corpora those models relied on, we skip the alignment problem entirely: we train directly on robot demonstrations from both sources and simply unify their action and state spaces using the general action representation described in Sec.~\ref{sec:action_repr}.

\paragraph{Dataset Staging.} Each training stage serves a different purpose, so each should be trained on a different data distribution. For midtraining, we require vast and highly diverse data sources, so we use the entire dataset. For posttraining, the model has already seen the full corpus and the stage mainly aims to improve long-horizon and real-time responsiveness rather than to broaden coverage. We therefore train it on small datasets collected for specific downstream tasks, which are long enough to contain complete episodes and representative of what the deployed simulator is asked to continue, rather than on a subsample of the pretraining mixture.

\subsection{Data Pipeline}

\subsubsection{Stage 1: Video Quality Filtering}

\paragraph{Spatiotemporal validity.}
We discard clips whose motion or temporal characteristics fall outside the
corpus distribution. Motion is assessed via dense optical flow~\cite{optical-flow}:
given the flow field $\mathbf{u}_t(p)$ between adjacent frames, we compute
\begin{equation}
  M(V) = \frac{1}{T-1} \sum_{t=1}^{T-1} \frac{1}{|\Omega|} \sum_{p \in \Omega} \|\mathbf{u}_t(p)\|_2,
  \label{eq:motion_score}
\end{equation}
where $\Omega$ is the image domain. The motion score is not used to retain only
high-motion samples; instead, it supports distribution control. Near-static
clips are downsampled so the corpus is not dominated by a small number of
platforms that remain stationary. Motion scores are read within a platform rather
than across the corpus, because the flow a camera records at rest varies markedly
between fixed, handheld and wrist-mounted views. The retained
data therefore spans subtle in-hand manipulation, camera movement, and larger
end-effector displacement without collapsing toward either motion extreme.

\paragraph{Artifact detection.}
We remove frames and clips containing JPEG encoding artifacts, and discard any
whose state or action values are numerically corrupted.

\paragraph{Episode length filtering.}
Clips shorter than 1 second are excluded, as they do not provide enough
temporal context for autoregressive training.

\begin{figure*}[!t]
  \centering
  \includegraphics[width=\textwidth]{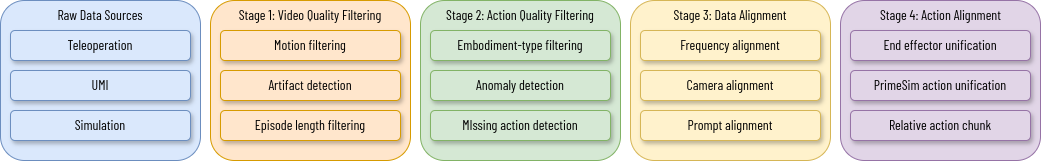}
  \caption{Overview of the data pipeline. Raw demonstrations from
  teleoperation, UMI, and simulation are converted through four stages:
  video quality filtering, action quality filtering, action preprocessing,
  and data alignment. Each stage lists the operations applied to the incoming
  data. The pipeline produces a unified corpus of 30,110 hours across 20
  dataset families and 22 robot platforms
  (Table~\ref{tab:datasets_frames}).}
  \label{fig:pipeline}
\end{figure*}

\subsubsection{Stage 2: Action Quality Filtering}

\begin{figure*}[!t]
  \centering
  \includegraphics[width=0.85\textwidth]{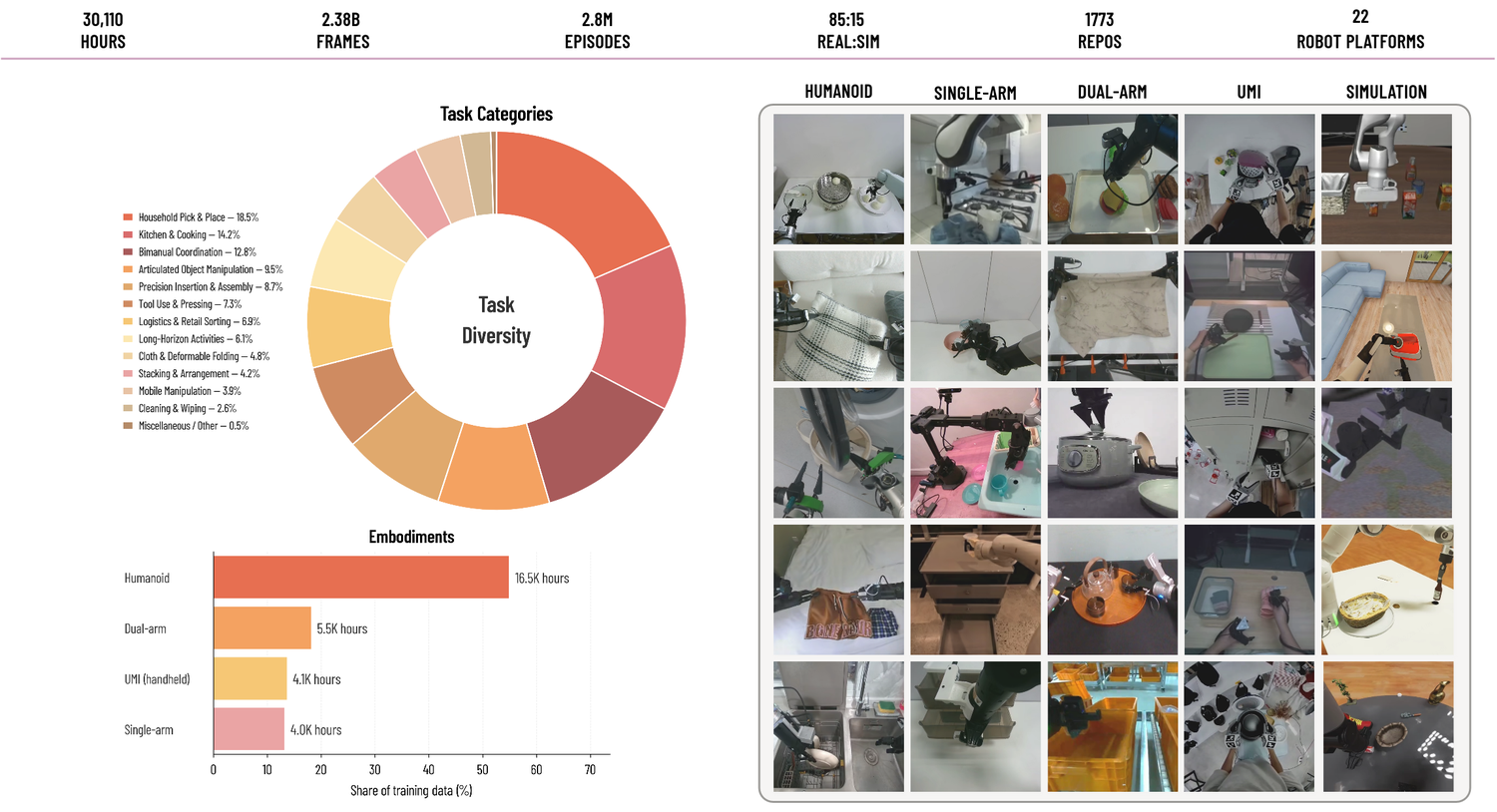}
  \caption{Overview of the PrimeSim training corpus: 30{,}110 hours, 2.38B
  frames and 1.52M episodes from 1{,}773 repositories over 22 robot platforms, at
  an 84:16 real-to-simulated ratio. \textbf{Left:} the task-category and
  embodiment distributions. \textbf{Right:} example frames from each of the
  five embodiment classes.}
  \label{fig:overview}
\end{figure*}

\paragraph{Embodiment-type filtering.}
Datasets with dexterous hands or humanoid legs represent a small fraction of the
pooled corpus. To keep the action representation simple, we filter these datasets
out. This removes legged locomotion, not humanoids as a class: humanoid platforms
on a wheeled chassis are kept, because their arm, torso and chassis commands all
fit the slots the interface already provides, and they supply the largest single
share of the corpus (Fig.~\ref{fig:overview}).

\paragraph{Anomaly detection.}
We discard sequences whose action signals contain discontinuities inconsistent
with continuous robot motion. We detect these as outliers in the first difference
of each action channel: an isolated step change far outside the distribution of
adjacent steps indicates a sensor or logging fault rather than a commanded
acceleration. Datasets containing corrupted video frames, or action values outside
their admissible range, are discarded as well.

\paragraph{Missing action detection.}
Some datasets leave certain action features blank, such as end-effector pose or
chassis velocity. If these values cannot be recovered through forward
kinematics, the affected data is dropped.

\subsubsection{Stage 3: Data Alignment}

\paragraph{Frequency alignment.}
We preserve each dataset's native sampling rate at collection time, but resample
to a fixed 10\,Hz during training. If a dataset cannot be resampled to exactly
10\,Hz within a 0.03\,s timestamp tolerance, we use the closest achievable
frequency.

\paragraph{Camera alignment.}
The number of cameras varies across embodiments. To make batches collatable, we
fix the model's camera count to three. Datasets with more than three cameras
have their uncommon cameras dropped; datasets with fewer have the primary
camera repeated until three slots are filled.

\paragraph{Prompt alignment.}
Datasets differ in whether they provide natural-language task descriptions. For
datasets that include prompts, we keep them as is. For datasets that do not, we
  generate a prompt from the task metadata with a template-based generator, so that
  every training sample is paired
  with a task description.

\paragraph{Format conversion.}
All datasets are converted to LeRobot~\cite{lerobot} format. We chose this format
for three reasons: (i) its storage is space-efficient, (ii) many open-source
datasets already use it, and (iii) recent updates have made it reliable for
large-scale training at the tens-of-thousands-of-hours scale used in this
project.

\subsubsection{Stage 4: Action Alignment}

\paragraph{End-effector unification.}
Datasets store end-effector pose inconsistently, and we unify it in three steps.
For datasets that provide joint-space commands without end-effector poses, we
look up the robot's open-sourced URDF and compute them using the forward
kinematics implementation in LeRobot~\cite{lerobot}. For datasets with poorly
defined action labels, we discard the labels and use proprioceptive data
instead, taking the pose one timestep ahead as the current timestep's action.
Rotations vary across datasets in both parametrization and quaternion component
ordering, so we convert every rotation to a matrix and then to a quaternion in
$(x, y, z, w)$ order, with the real value last. Datasets lacking both
end-effector poses and usable joint-space commands are excluded.

\paragraph{PrimeSim action unification.}
To unify actions across heterogeneous embodiment platforms, we convert every
action into a shared representation consisting of bimanual end-effector poses
with quaternion rotations, full-body joint angles, gripper commands, and chassis
velocity in $(v_x, v_y, \omega)$. Gripper commands are binary open/close signals
and are kept in absolute form. We refer the reader to
Sec.~\ref{sec:action_repr} for the full specification.

\paragraph{Relative action chunk.}
We convert absolute actions into relative action chunks by expressing every
action in the coordinate frame of the proprioceptive state at frame~0 of the
current window. For body joints, we take the difference between the current
joint configuration and the reference configuration,
\begin{equation}
  \tilde{q}_t = q_t - q_0 .
  \label{eq:rel_joints}
\end{equation}
For end-effector pose, we express the current translation and rotation in the
reference frame. Let $p_t \in \mathbb{R}^3$ and $R_t \in SO(3)$ denote the
end-effector translation and rotation at time $t$, and let $p_0, R_0$ denote
their values at the reference frame. The relative pose is
\begin{align}
  \tilde{p}_t &= R_0^{\top} (p_t - p_0), \label{eq:rel_pos} \\
  \tilde{R}_t &= R_0^{\top} R_t . \label{eq:rel_rot}
\end{align}
Equivalently, in homogeneous form, $\tilde{T}_t = T_0^{-1} T_t$ where
$T_t = [R_t \mid p_t]$. Gripper commands and chassis velocity are left
unchanged, as both are already expressed in a frame-independent form.

\begin{figure*}[!t]
    \centering
    \includegraphics[width=1.0\textwidth]{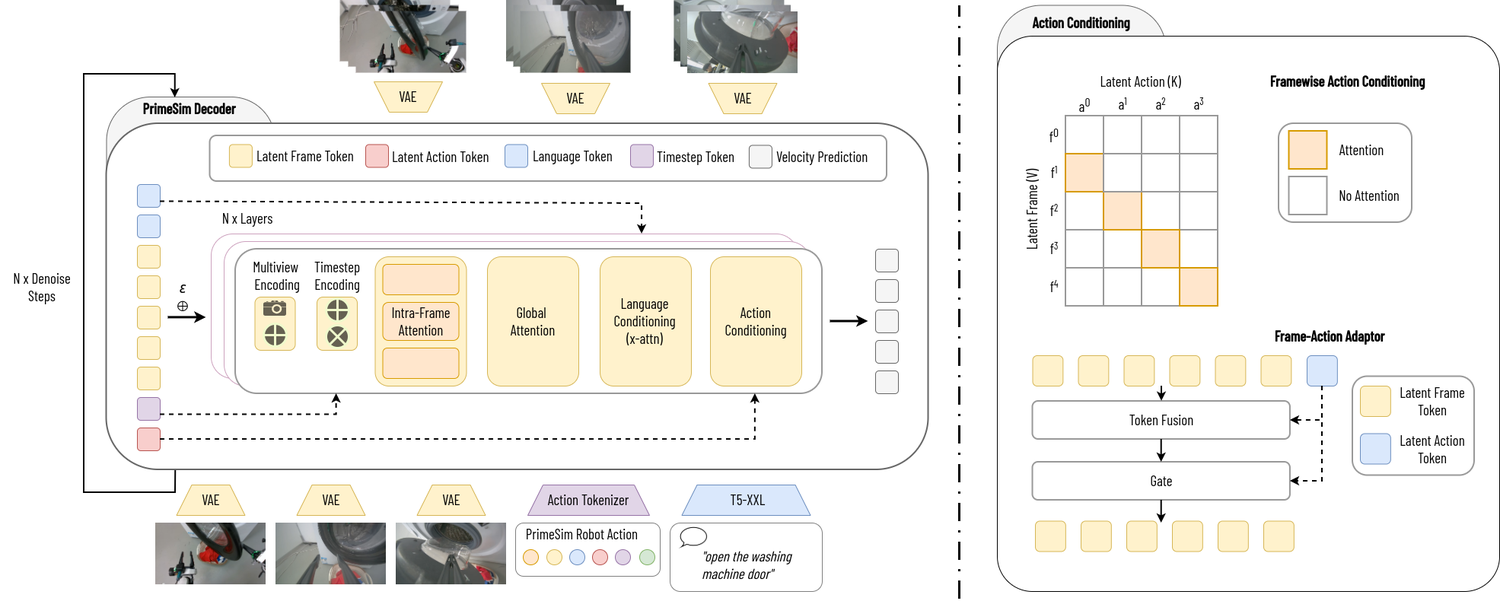}
    \caption{The PrimeSim model architecture. \textbf{Left:} the whole system,
    including the input data modalities, the VAE, the text encoder and the
    PrimeSim decoder. \textbf{Right:} the action conditioning mechanisms, a
    framewise attention mask and a frame-action adaptor.}
    \label{fig:architecture}
\end{figure*}
\section{PrimeSim}
\label{sec:primesim}
PrimeSim aims to serve as a world model simulator for long-horizon
robot manipulation, generating future camera observations
conditioned on robot actions. Unlike conventional video
generation, robot simulation requires the model to faithfully
execute prescribed actions, maintain consistency across
camera views, and preserve coherent scene dynamics over
extended trajectories. Moreover, the simulator must support
low-latency autoregressive generation to enable efficient
interaction with robot policies.

To address these requirements, we adopt a progressive
training framework that transforms a pretrained video
generator into an action-conditioned, multiview,
autoregressive simulator. We first initialize PrimeSim
from a pretrained video generation backbone that provides
general visual and motion priors. During \emph{midtraining},
we introduce structured action conditioning and multiview
adaptation to learn robot dynamics from large-scale
demonstrations. During \emph{posttraining}, we progressively
adapt the model for causal generation, few-step sampling,
and long-horizon self-conditioned rollouts with persistent
temporal memory.

\subsection{Pretrain}
\label{sec:pretrain}

We initialize from the Wan2.2-TI2V-5B pretrained weights \cite{wan}, which
provides a strong generative prior over natural video: object permanence, lighting
and material statistics, and short-horizon motion patterns that would be
prohibitively expensive to relearn from robot data alone. This stage establishes
the general world priors, spatiotemporal coherence and high-fidelity texture that
the later stages build on. We adopt this checkpoint as given: all of our own
training happens in the midtraining and posttraining stages that follow.

\subsection{Midtrain}
\label{sec:midtrain}
Midtraining adapts the pretrained video generator to
action-conditioned robot dynamics. While the pretrained
backbone provides general visual and motion priors, it lacks
explicit control over robot actions and consistent generation
across multiple camera views. To address these limitations,
we introduce three complementary components: a unified
cross-embodiment action representation, frame-aligned dual-path
    conditioning, and view-aware interleaved
attention. We jointly train these components with the
video generation backbone on 30{,}110 hours of robot
demonstrations spanning 22 platforms
(Sec.~\ref{sec:data_engine}).

\paragraph{Problem Formulation.}
Our objective during midtraining is to learn the visual
consequences of prescribed robot actions across multiple
camera views. At robot timestep $t$, let
$o_t=(\mathbf{I}_t,q_t)$ denote the observation, where
$\mathbf{I}_t=(I_t^1,\ldots,I_t^n)$ contains images
from $n$ cameras and $q_t$ represents the robot's
proprioceptive state.

Given an observation context
$\mathcal{O}_t=(o_{t-K+1},\ldots,o_t)$,
an $H$-step action trajectory
$A_t=(a_{t+1},\ldots,a_{t+H})$,
and an optional language instruction $\ell$,
we formulate multiview video prediction as
\begin{equation}
\mathbf{I}_{t+1:t+H}
\sim
\mathcal{W}_{\mathrm{BD}}
(\cdot\mid\mathcal{O}_t,A_t,\ell),
\label{eq:midtrain_formulation}
\end{equation}
where $\mathcal{W}_{\mathrm{BD}}$ denotes the
bidirectional world model learned during midtraining.
Unlike autoregressive simulation, the model jointly
predicts the future video within a fixed temporal
window. This formulation provides the foundation
for the action and multiview conditioning mechanisms
described below.

\subsubsection{Action Representation}
\label{sec:action_repr}

\begin{wrapfigure}{r}{0.38\columnwidth}
  \centering
  \vspace{-2.0\baselineskip}
  \captionsetup{font=footnotesize, justification=raggedright}
  \includegraphics[width=\linewidth]{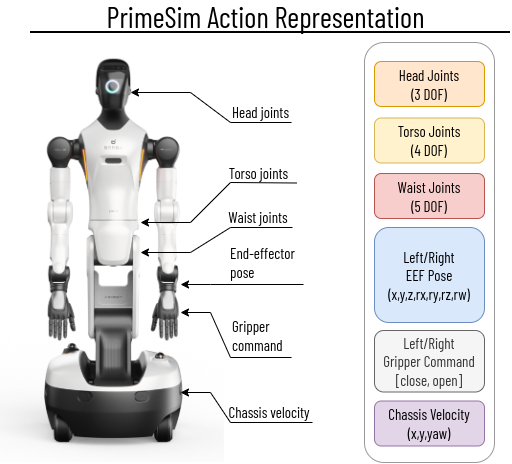}
  \caption{General action representation in PrimeSim, covering single-arm,
  bimanual, UMI, and wheeled humanoid embodiments.}
  \label{fig:action_repr}
  \vspace{-0.8\baselineskip}
\end{wrapfigure}
We define actions as changes in robot poses. Given consecutive video tokens, an
action token $a_t$ denotes the pose change that produces the transition into
latent frame $t$. Each embodiment source is transformed into a fixed
31-dimensional representation that captures a shared underlying geometric
structure across different action domains, with the slots an embodiment lacks set
to zero.

\paragraph{General Action Representation.} At a high level, actions can include up to four components: effector poses for
the agent's effectors, grasp states, pose of the head, waist, and torso, and the chassis velocity. To cover manipulation poses across almost all robot platforms' pose representation, we represent arm states as end-effector poses, grasp states as binary
flags, head/torso/waist as joint angles, and the chassis as a planar velocity
$(v_x, v_y, \omega)$. As each embodiment's degrees of freedom varies, we leave redundancy in each joint slot to account for this (see Fig.~\ref{fig:action_repr}).

\paragraph{Relative Action Chunk.} All actions are defined relative to the initial frame of each window, i.e., $\tilde{a}_{t+i\mid t} = a_{t+i} \ominus a_t$. This relative action representation for conditioning resolves several core issues in cross-embodiment learning. First, embodiments with missing features can simply be represented as $0$ (no movement). Second, the absolute origin offset of each embodiment is cancelled out by the relative transformation. Third, it strengthens the per-frame action conditioning signal by grounding all changes to the conditioning first frame, as well as concentrating the action distribution. Relative actions have been explored in prior models, but with deltas taken from the previous frame \cite{dreamdojo}. We argue that anchoring to the first conditional frame of the window instead provides a stronger conditioning signal for world models. An ablation for relative action chunk is provided in Table~\ref{tab:relative_action}.

\subsubsection{Action Conditioning}

Effectively controlling a world model to generate future frames
that faithfully follow commanded actions remains a fundamental
challenge. Existing models often hallucinate robot motions
inconsistent with the conditioned actions~\cite{abot-physworld,bwm}.
We attribute this limitation in part to the lack of explicit
temporal alignment between actions and generated frames.
Under standard global attention, each frame can aggregate
action information from multiple timesteps, making it
difficult to identify which action should govern its
evolution. To address this, we introduce a
\emph{frame-aligned dual-path conditioning} mechanism, which
strengthens action--frame correspondence by explicitly binding
actions to their corresponding frames through two complementary
pathways.
while the second directly injects action information into
frame representations, ensuring that action conditioning does
not rely solely on attention routing.
Together with our first-frame-anchored relative action
representation, these designs provide temporally precise
and explicit control over generated robot motions.
We evaluate their individual contributions in
Table~\ref{tab:action_cond_ablation}.

\paragraph{Action Tokenizer.}
To enable frame-aligned conditioning, we first establish
a temporal correspondence between robot actions and
video latents. Specifically, we encode each relative
action chunk into a compact sequence of latent action
tokens using a two-layer MLP, producing one action token
for each latent video frame.
This representation provides a shared temporal structure
for the two conditioning pathways described below,
allowing each frame to be associated with the action
responsible for its evolution.

\paragraph{Framewise Action-Conditioning.}
Given the aligned action tokens, our first conditioning
pathway explicitly constrains how actions interact with
video representations.
Rather than allowing every frame to attend to actions
from arbitrary timesteps, we introduce a shifted-diagonal
attention mask that restricts frame-to-action attention
to the corresponding temporal transition.
Specifically, since action $a_t$ drives the transition
into frame $t+1$, frame $t$ attends only to $a_{t-1}$,
while the initial conditioning frame receives no action
conditioning. This prevents interference from temporally
unrelated actions and establishes an explicit
action--frame correspondence at the attention level.
While prior work has explored frame-level action
conditioning~\cite{ctrl-world}, our approach enforces
this correspondence directly through the attention mask
and combines it with first-frame-anchored relative
action chunks, which specify the intended motion
relative to a consistent reference state.

\paragraph{Action-Frame Adapter.}
Although framewise attention establishes which action
each frame should attend to, it does not guarantee
that the corresponding action signal sufficiently
influences the frame representation.
In particular, action information may still be
underweighted when competing with rich visual features
during attention.
We therefore introduce a complementary conditioning
pathway that directly injects the aligned action
information into each frame token.
Specifically, a lightweight action--frame adapter
computes a residual feature shift from the corresponding
action token and integrates it into the frame
representation (Figure~\ref{fig:architecture}).
Unlike attention-based conditioning, this pathway
provides a direct route for action information,
independent of the learned attention weights.
The two pathways thus serve distinct but complementary
roles: framewise attention determines \emph{which}
action governs each frame, while the adapter ensures
that the selected action directly modulates its
representation. Their combination strengthens
action--frame binding and encourages generated
motions to remain consistent with the specified
action trajectory.

\subsubsection{Multiview Adaption}
A fundamental challenge in action-conditioned multiview generation is ensuring that different camera views remain consistent with the same evolving physical scene. This is particularly challenging in robotics, where wrist-mounted cameras move with the robot and may observe one another, creating strong dependencies across views.
Treating each view independently can lead to inconsistent
robot motions and object interactions, while unrestricted
global attention may obscure view-specific information.
To address this, we introduce \emph{view-aware interleaved attention}
with explicit camera identities, which enables cross-view information
exchange while preserving view-specific representations.

Specifically, we organize the 30 DiT blocks to alternate
between intra-view and global attention: two out of every
three blocks restrict attention to tokens within each
camera view, while every third block enables information
exchange across all three views. This design allows the
model to capture view-specific spatial and temporal
relationships locally, while periodically integrating
information across cameras to maintain a coherent
representation of the shared physical scene.

However, effective cross-view interaction also requires
the model to distinguish the source of each visual token,
particularly when features from different cameras are
mixed through global attention. We therefore prepend a
dedicated camera token to each view, explicitly encoding
its view identity throughout the generation process.
The two designs serve complementary roles:
interleaved attention controls \emph{how} information
is exchanged across views, while camera tokens preserve
\emph{which} view each representation belongs to.
Together, they promote consistent multiview generation
without sacrificing view-specific information
(Figure~\ref{fig:architecture}).
We validate the contribution of both components in
Table~\ref{tab:multiview_ablation}.

\subsubsection{Training Objective}
\label{sec:midtrain_objective}

Having established the unified action representation,
frame-aligned action conditioning, and view-aware
multiview attention, we jointly optimize these
components under a conditional latent flow-matching
objective. Rather than introducing separate objectives
for action following or multiview consistency, both
capabilities are learned through the conditional
generation of future multiview observations.

Let
$x_0=\mathcal{E}(\mathbf{I}_{t+1:t+H})$
denote the clean latent representation of the target
multiview video, where $\mathcal{E}$ is the video
encoder. We collect the conditioning inputs as
$c=(\mathcal{O}_t,A_t,\ell)$, comprising the
observation context, action trajectory, and optional
language instruction.

For a sampled flow timestep $\tau\in[0,1]$ and
Gaussian noise $\epsilon\sim\mathcal{N}(0,\mathbf{I})$,
we construct the noisy latent and its target velocity:
\begin{equation}
x_\tau=(1-\tau)x_0+\tau\epsilon,
\qquad
v_\tau=\epsilon-x_0.
\label{eq:flow_path}
\end{equation}

The conditional velocity network $u_{\theta_{\mathrm{BD}}}$
is trained to predict the target velocity:
\begin{equation}
\mathcal{L}_{\mathrm{flow}}(\theta_{\mathrm{BD}})
=
\mathbb{E}_{(x_0,c),\epsilon,\tau}
\left[
\left\|
u_{\theta_{\mathrm{BD}}}(x_\tau,\tau,c)
-v_\tau
\right\|_2^2
\right],
\label{eq:flow_matching}
\end{equation}
where $\theta_{\mathrm{BD}}$ denotes the parameters
of the bidirectional world model, and $\tau$
indexes the generative flow process rather than
robot time.

Importantly, all camera views are generated jointly,
and robot actions are incorporated directly into the
velocity prediction network through the proposed
conditioning mechanisms. This allows the model to
learn action-conditioned visual dynamics and
cross-view dependencies under a unified generative
objective.

The resulting bidirectional world model
$\mathcal{W}_{\mathrm{BD}}$ provides the foundation
for subsequent autoregressive adaptation.
During posttraining, we retain flow matching for
initial causal adaptation, while introducing
consistency distillation and distribution matching
to enable efficient, self-conditioned, long-horizon
simulation.

\clearpage
\subsection{Posttrain}
\label{sec:posttrain}

\begin{figure*}[!t]
  \centering
  \includegraphics[width=0.7\textwidth]{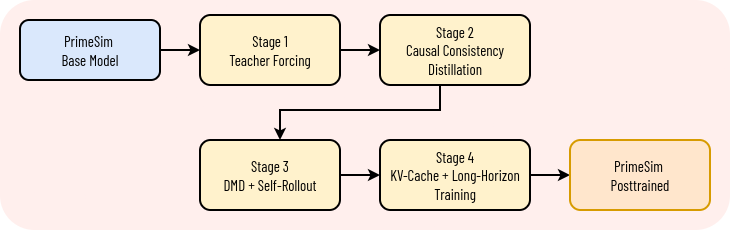}
  \caption{Overview of the post-training pipeline, from the midtrained
  base model through the four adaptation stages to the deployed simulator.}
  \label{fig:posttrain_pipeline}
\end{figure*}

Midtraining equips PrimeSim with the ability to predict action-conditioned,
multiview robot videos within a fixed temporal window. However, accurate
short-horizon generation does not by itself yield a practical world model simulator.
The midtrained model generates clips with bidirectional attention and requires
many denoising steps, whereas policy interaction calls for continuously
extending a rollout from the model's own predictions. This introduces three
additional challenges: autoregressive generation must respect temporal
causality, sampling must be fast enough for repeated rollouts, and errors and
missing scene context must not accumulate uncontrollably over long horizons.

We therefore posttrain PrimeSim to transform \emph{fixed-window video
prediction} into \emph{fast, self-conditioned, long-horizon simulation}. Our
guiding principle is \emph{fast video, slow memory}: generate each new chunk
with few-step sampling while retaining informative history across much longer
timescales. We realize this through four progressive stages: (1) causal
adaptation enables chunk-wise autoregressive generation; (2) consistency
distillation produces a four-step sampler; (3) clip-level self-rollout with
asymmetric distribution matching improves generation conditioned on the
model's own outputs; and (4) episode-level adaptation combines self-rollout
training with bounded, multiscale temporal memory. The stages are sequential:
causal generation enables rollout, fast sampling makes rollout training
tractable, and self-conditioned training prepares the model for long-horizon
continuation.

\subsubsection{Stage 1: Autoregressive Modelling}

To enable continuous simulation, we first adapt the bidirectional model
$\mathcal{W}_{\mathrm{BD}}$ obtained from midtraining into a
chunk-causal model $\mathcal{W}_{\mathrm{C}}$.
Unlike the midtrained model, which jointly generates all frames
within a fixed temporal window, the causal model generates video
chunks sequentially, conditioned on previously generated observations.
We achieve this through \emph{chunk-causal attention}, which
preserves bidirectional attention within each chunk while restricting
cross-chunk attention to preceding chunks.
This allows the model to retain its learned action-conditioned
multiview generation capabilities while supporting autoregressive
prediction.

Specifically, we adopt teacher forcing
~\cite{self-forcing,self-forcing++,diffusion-forcing,causal-forcing,longlive}
during causal adaptation, where each chunk is conditioned on clean
ground-truth history. Let $B_{i,\mathrm{clean}}$ and
$B_{i,\mathrm{noisy}}$ denote the clean and noisy representations
of chunk $i$, respectively.
The attention context for each noisy chunk is defined as
\begin{equation}
\mathcal{S}\!\left(B_{i,\mathrm{noisy}}\right)
=
\left\{B_{j,\mathrm{clean}}:j<i\right\}
\cup
\left\{B_{i,\mathrm{noisy}}\right\},
\label{eq:chunk_causal_mask}
\end{equation}
where $\mathcal{S}(\cdot)$ denotes the accessible attention
context. This design prevents information leakage from future
chunks while allowing all chunks to be trained in parallel.
The same chunk partition also determines the noise granularity
of the diffusion-forcing objective. Stage~1 therefore keeps the flow-matching
loss of Eq.~\ref{eq:flow_matching}; what changes is the attention mask the loss
is evaluated under, which is what we mean by diffusion forcing rather than a new
objective.

Through this adaptation, PrimeSim acquires the ability to
generate future observations autoregressively rather than
within a fixed bidirectional window.
However, teacher forcing introduces an exposure bias:
the model is trained with clean historical observations
but must condition on its own imperfect predictions at inference.
Increasing the chunk size from $k=1$ to $k=3$ reduces the number
of sequential generation chunks from nine to three in our
training window, mitigating but not eliminating error accumulation.
To enable subsequent training on self-generated histories,
we first reduce the computational cost of autoregressive
generation through few-step distillation in Stage~2.

\subsubsection{Stage 2: Causal Consistency Distillation}
\label{sec:cd}

Although Stage~1 enables autoregressive generation, the resulting
causal model $\mathcal{W}_{\mathrm{C}}$ still requires 50 denoising
steps per chunk, making continuous simulation computationally
expensive. More importantly, the high sampling cost makes it
impractical to train on self-generated rollouts, where gradients
must propagate through multiple generation chunks.
To address this, we adopt Causal Consistency Distillation
(Causal CD)~\cite{causal-forcing++} to obtain an efficient
four-step generator. Specifically, we use $\mathcal{W}_{\mathrm{C}}$
as a frozen \emph{causal teacher} and initialize a
\emph{causal student} $\mathcal{W}_{\theta}$ from its weights.
Both models share the same chunk-causal architecture, allowing
the student to preserve the autoregressive structure learned
in Stage~1 while substantially reducing the number of
sampling steps.

Rather than supervising the student with complete
teacher-generated trajectories, Causal CD performs online
distillation by matching local transitions of the teacher's
flow trajectory. Let $x_t$ denote a noisy video latent at
flow time $t$, $\hat{x}_{0,\mathrm{C}}$ the teacher's clean
estimate, and $x_{t-\Delta t}^{\mathrm{C}}$ the target
obtained from a local teacher ODE step.
The distillation objective is
\begin{equation}
\mathcal{L}_{\mathrm{CD}}(\theta)
=
\mathbb{E}_{x_t,t,\Delta t}
\left[
w(t)\,
d\!\left(
G_{\theta}(x_t,\hat{x}_{0,\mathrm{C}},t),
x_{t-\Delta t}^{\mathrm{C}}
\right)
\right],
\label{eq:cd}
\end{equation}
where $G_{\theta}$ denotes the student's local flow-map
update and $d(\cdot,\cdot)$ measures the discrepancy
between student and teacher transitions.
By matching nearby flow states, the student learns
to approximate the teacher's multi-step generation
process with substantially fewer sampling steps.
Moreover, the online objective requires only a local
teacher ODE step per update, avoiding the expensive
precomputation and storage of complete teacher
trajectories.

To further reduce the memory overhead of maintaining
teacher and student models simultaneously, we optimize
LoRA~\cite{lora} adapters while keeping the midtrained
backbone weights frozen. Together, these designs produce
a four-step causal student that substantially accelerates
autoregressive generation and, crucially, makes
self-rollout training computationally tractable.
Nevertheless, distillation alone does not resolve the
exposure bias inherited from teacher-forced training,
nor does it guarantee visual fidelity under repeated
self-conditioned generation. We therefore further
optimize the distilled student on its own generated
rollouts in Stage~3.

\subsubsection{Stage 3: Clip-Level Self-Rollout with Asymmetric DMD}
\label{sec:dmd}

Although Stage~2 substantially accelerates autoregressive generation,
the distilled student $\mathcal{W}_{\theta}$ still faces two
limitations. First, training with clean historical observations
introduces exposure bias, as the model must condition on its own
imperfect predictions during inference. Second, consistency distillation 
has a forward-KL-like, mode-covering effect, potentially
producing averaged predictions with reduced perceptual detail.
To address both limitations, we introduce \emph{clip-level
self-rollout training with asymmetric distribution matching}.
Specifically, the student autoregressively generates video chunks
within each training clip, with subsequent chunks conditioned on
its own previously generated latents through the causal KV cache,
rather than on ground-truth history.
This exposes the model to the conditioning distribution
encountered during inference.

To preserve generation quality under self-conditioned rollouts,
we further optimize the student using an asymmetric Distribution
Matching Distillation (DMD) objective~\cite{dmd,dmd2}.
Unlike Stage~2, which distills a causal teacher into a causal
student, we now use the frozen bidirectional model
$\mathcal{W}_{\mathrm{BD}}$ from midtraining as the reference
teacher. This design exploits the stronger fixed-window
generation capability of the bidirectional model to supervise
the autoregressive student, without imposing the teacher's
bidirectional attention structure on the student.
We adopt the mode-seeking reverse-KL objective
\begin{equation}
D_{\mathrm{KL}}\left(
p_{\theta}(\cdot\mid c)
\,\|\,
p_{\mathrm{BD}}(\cdot\mid c)
\right),
\end{equation}
where $p_{\theta}$ and $p_{\mathrm{BD}}$ denote the
conditional distributions of the causal student and
bidirectional reference model, respectively, and $c$
represents the corresponding conditioning information,
including visual context, robot actions, and camera-view
identities.

To optimize this objective, we introduce a trainable
\emph{fake-score network} $v_{\phi}$ that estimates
the distribution of student-generated samples.
Together with the frozen bidirectional reference, it
provides the score difference used to update the student.
Let $\tilde{x}_0$ denote the clean video latent generated
by the student, and let $\hat{x}_{0,\phi}$ and
$\hat{x}_{0,\mathrm{BD}}$ denote the clean-sample estimates
of the fake-score network and bidirectional teacher,
respectively, evaluated on the same noisy latent.
We construct the normalized score-difference direction as
\begin{equation}
\nabla_{\mathrm{KL}}
=
\frac{
\hat{x}_{0,\phi}-\hat{x}_{0,\mathrm{BD}}
}{
\max\!\left(
\mathbb{E}\left|
\tilde{x}_0-\hat{x}_{0,\mathrm{BD}}
\right|,
\delta
\right)
},
\qquad \delta=10^{-8},
\label{eq:kl_grad}
\end{equation}
where the normalization stabilizes the update when the
student's generated distribution deviates substantially
from the reference.
The student is optimized using the surrogate loss
\begin{equation}
\mathcal{L}_{\mathrm{gen}}(\theta)
=
\frac{1}{2}
\left\|
\tilde{x}_0-
\operatorname{sg}\left(
\tilde{x}_0-\nabla_{\mathrm{KL}}
\right)
\right\|_2^2,
\label{eq:gen_surrogate}
\end{equation}
where $\operatorname{sg}$ denotes the stop-gradient
operator. Meanwhile, the fake-score network is trained
on student-generated samples using the standard flow
matching objective,
\begin{equation}
\mathcal{L}_{\mathrm{critic}}(\phi)
=
\mathbb{E}_{\tilde{x}_0,\epsilon,\tau}
\left[
\left\|
v_{\phi}(x_{\tau},\tau,c)
-
(\epsilon-\tilde{x}_0)
\right\|_2^2
\right],
\label{eq:critic}
\end{equation}
where
$x_{\tau}=(1-\tau)\tilde{x}_0+\tau\epsilon$
is the noisy interpolation of the generated latent,
and $\epsilon\sim\mathcal{N}(0,I)$.

Through clip-level self-rollout and asymmetric distribution
matching, Stage~3 enables the four-step causal student to
learn from its own generated history while receiving
high-quality supervision from the bidirectional reference.
However, training remains confined to fixed-length clips,
whereas practical robot simulation requires coherent
generation over complete episodes.
This remaining short-training--long-inference gap motivates
the episode-level self-rollout and persistent temporal
memory introduced in Stage~4.

\subsubsection{Stage 4: Episode-Level Long-Horizon Adaptation}
\label{sec:selfrollout}

While Stage~3 enables self-conditioned generation within short
clips, reliable robot simulation requires maintaining coherent
predictions over trajectories far longer than the training
window. At this scale, two challenges become particularly
critical: prediction errors accumulate as generated chunks
repeatedly serve as future conditioning, and important scene
information may be lost from recent context due to occlusion
or camera motion. To address both challenges, we extend
self-rollout training from fixed-length clips to complete
robot episodes and introduce a \emph{persistent hierarchical
temporal memory}. The former exposes the model to long-horizon
error accumulation during training, while the latter preserves
scene information across multiple temporal scales.
Together, they enable the student to generate extended
trajectories using its own predictions rather than relying
on clean historical observations.

Stage~4 inherits the asymmetric DMD objective of Stage~3 unchanged. What
changes is what the rollouts span: the student is now supervised on episodes
rather than fixed clips, so the distribution it is matched against includes the
errors it accumulates over a long horizon rather than only over a few chunks.

\paragraph{Episode-level self-rollout.}
We first extend self-conditioned generation to complete
robot episodes, allowing the student to experience the
long-horizon conditioning distribution encountered during
inference. Specifically, the student generates successive
video chunks autoregressively, with each continuation
conditioned on its own previously generated history through
the causal KV cache. Generator gradients are accumulated
across the episode, followed by a single optimization
update at the episode boundary. Meanwhile, the fake-score
network is updated after each generated window to track
the evolving student distribution. The KV cache persists
throughout the episode and is reset only when a new
episode begins. Unlike clip-level training, this procedure
directly exposes the student to errors that accumulate
across successive generations, encouraging more stable
long-horizon predictions.

Memory in world models is usually conditioned on encoded ground-truth
observations~\cite{ctrl-world, geniesim2, genie-envisioner, bwm} or on a cache of
clean historical frames~\cite{memorywam, context-as-memory, memorypack}. Both
supply, at inference, data the model does not itself produce, which opens a
train-inference gap. PrimeSim instead trains its memory on its own rollouts, so
the distribution it conditions on during training is the one it will have at
inference.

\paragraph{Hierarchical temporal memory.}
Although episode-level self-rollout exposes the model to
longer trajectories, it does not by itself prevent the
loss of historical information. Retaining every generated
frame is computationally expensive, while keeping only
recent frames can discard important scene states.
We therefore introduce a \emph{hierarchical temporal
memory} that organizes the KV cache into three
complementary components: \emph{global sink},
\emph{slow memory}, and \emph{local memory}.

The global sink retains the initial observed latent state
as a persistent reference throughout the episode.
The local memory stores recent consecutive generated
latents, preserving fine-grained motion information.
The slow memory, in contrast, maintains a sparsely
sampled representation of longer-term history.
After each generation window, the previous local-memory
entries are merged with the existing slow-memory bank
and temporally downsampled to a fixed capacity, while
the newly generated latents populate the next local-memory
bank. Temporal positions are preserved during downsampling,
and synchronized frame indices are used across all camera
views to maintain aligned multiview histories.

This hierarchical design allows each generated chunk
to access the initial scene, sparse long-term context,
and dense recent motion within a bounded memory budget.
Importantly, except for the initial observed state,
all memory entries originate from the student's own
generated history. The same memory update mechanism
is used during training and inference, avoiding the
distribution mismatch introduced by teacher-forced
historical context.

\paragraph{Latent-space continuation.}
To further improve the efficiency and consistency of
long-horizon generation, we maintain the student's
continuation entirely in latent space. Rather than
decoding each generated segment and re-encoding it
as a new conditioning observation, the student carries
its history forward through the persistent KV cache.
Continuation windows receive neither a new first-frame
latent nor a clean ground-truth prefix, ensuring that
the autoregressive process remains conditioned on
the model's own predictions. This avoids repeated
video decoding and encoding along the generation
path and allows the student to extend trajectories
without repeatedly reinitializing visual context.

\paragraph{Amortized supervision.}
A remaining challenge is supervising episode-length
student rollouts using the bidirectional reference
and fake-score networks, which operate within fixed
temporal windows. We decouple long-horizon generation
from local distribution supervision. For each continuation,
we decode the preceding segment once and re-encode only
its final frame as a conditioning observation for the
score networks, as required by the causal VAE.
This operation is performed outside the student's
gradient path, and the re-encoded observation is never
fed back into the student's autoregressive generation.
Consequently, the student can accumulate history across
complete episodes, while the score networks evaluate
local transitions within their original temporal windows.
This enables long-horizon self-rollout training without
extending the temporal context of the supervising models.

\paragraph{Action re-anchoring.}
Finally, to preserve consistent action conditioning
across long rollouts, we express actions relative to
the reference pose recorded at the beginning of each
episode, rather than resetting the reference at every
generation window. This maintains a consistent action
coordinate frame across continuation boundaries and
prevents changes in the reference frame from introducing
ambiguity into the conditioned motion trajectory.

Together, the four posttraining stages transform
PrimeSim from a fixed-window, action-conditioned
video generator into a \emph{fast, self-conditioned,
long-horizon world model simulator}. Few-step distillation
enables efficient autoregressive generation, while
episode-level self-rollout and hierarchical temporal
memory support coherent simulation across extended
robot trajectories. At inference, PrimeSim requires
only the four-step causal student and its bounded
KV cache, generating future multiview observations
conditioned on robot actions without access to
ground-truth future frames.

\section{Experiments}
\label{sec:experiments}

Six sets of experiments follow, in the order the paper's argument develops. We
first place PrimeSim against existing world models on benchmarks built for
embodied prediction, together with the inference cost that decides whether a model
can serve as a simulator at all. We then ask what midtraining actually acquires,
by adapting to platforms and to scenes and camera placements the model has never
seen. We examine what post-training changes over long horizons, whether the
simulator is faithful enough to evaluate and to improve a policy, and finally
which of the design choices carry the result.

\subsection{Comparison to the State of the Art}
\label{sec:sota}

We compare on three axes, chosen because a simulator has to be right about
different things than a video generator. \emph{Trajectory fidelity} asks whether
the commanded motion is actually realised; \emph{scene and semantic consistency}
asks whether the world still looks and means the same while that motion happens;
and \emph{inference cost} asks whether the model can be run inside a control loop
at all. WorldSimProbe separates the first two into its own axes, EWMBench
aggregates them into a single score, and neither measures the third, so we report
throughput separately.

\subsubsection{WorldSimProbe Evaluation}
\label{sec:worldsimprobe_exp}

WorldSimProbe \cite{worldsimprobe} isolates a property that reconstruction
metrics do not capture: whether a world model behaves as a simulator. It is
built on the Observable Simulator Contract, which requires that supplied actions
induce corresponding agent motion, and that the environment respond to that
realized motion rather than to the action that was merely commanded. This
targets a failure mode invisible to PSNR, in which the robot does not move as
instructed while the scene reacts as though it had.

The benchmark decomposes the observable simulator contract into five diagnostic suites. T1 (Local Action Calibration) tests whether the model reproduces the simulator’s response to small and large action perturbations. T2 (Global Trajectory Coverage) evaluates whether it follows globally different but physically executable trajectories transferred across tasks and scenes. T3 (Action-Source Behavior Preservation) examines whether behavioral differences among expert, human-teleoperated, and policy-generated actions, such as speed, smoothness, and corrective motion, are retained. T4 (Interaction Grounding) uses validated no-contact counterfactuals to test whether the model avoids hallucinating interactions unsupported by the realized robot motion. T5 (Interaction Dynamics) assesses whether both robot motion and object response match intended interaction primitives such as pushing, pulling, rotating, shaking, and dropping.

We evaluate PrimeSim with a sink token that provides lightweight historical context for bidirectional architectures. To improve physical understanding, we additionally construct training data from three sources: multiple trajectories sampled from VLA task executions, perturbed variants of these sampled trajectories, and trajectories obtained after randomly replacing objects in the scene. Random visual noise is also applied during training to reduce the distribution shift caused by the mismatch between training and evaluation resolutions. PrimeSim generates \(192\times256\) images per view and predicts 33 frames at 10 FPS. Because the benchmark does not explicitly score appearance, task success, or coarse rollout-level responsiveness, the resulting comparison is primarily behavioral, and high scores indicate simulator fidelity rather than visual quality or downstream utility.

\begin{table}[t]
  \centering
  \caption{WorldSimProbe results on LIBERO~\cite{libero}, against five open-source
  action-conditioned world models. T1--T3 measure action realization (local action
  calibration, global trajectory coverage, action-source behavior preservation);
  T4--T5 measure interaction response (interaction grounding, interaction dynamics).
  Bold marks the best value per suite; higher is better. PrimeSim leads four of the
  five, all but local action calibration.}
  \label{tab:worldsimprobe}
  \begin{tabular}{l c c c c c c}
    \toprule
    Model & T1 $\uparrow$ & T2 $\uparrow$ & T3 $\uparrow$ & T4 $\uparrow$ & T5 $\uparrow$ & Mean $\uparrow$ \\
    \midrule
    LingBot-VA      & \textbf{77.0} & 60.2 & 51.0 & 57.8 & 22.4 & 53.7 \\
    DreamDojo       & 69.5 & 41.5 & 50.3 & 65.3 & 23.5 & 50.0 \\
    Cosmos-3-Nano   & 67.3 & 55.9 & 47.4 & 61.1 & 25.0 & 51.3 \\
    IRASim          & 54.0 & 54.7 & 47.9 & 68.2 & 22.4 & 49.4 \\
    Ctrl-World      & 62.7 & 61.7 & 50.4 & 77.5 & 25.0 & 55.5 \\
    \midrule
    PrimeSim        & 70.2 & \textbf{67.9} & \textbf{72.3} & \textbf{85.7} & \textbf{37.1} & \textbf{66.6} \\
    \bottomrule
  \end{tabular}
\end{table}

Against five open-source action-conditioned world models, PrimeSim ranks first on
four of the five suites, leading two of the three action realization axes (T2 at 67.9 and
T3 at 72.3) and interaction dynamics (T5 at 37.1). The one axis we do not win
is informative rather than incidental. LingBot-VA leads local action
calibration (77.0 against 70.2), indicating that our fine-grained control
precision is not yet the best in the field even where coarse trajectory
coverage is. Interaction grounding (T4) is led by us at 85.7 against Ctrl-World's 77.5, and it
follows from the same design decision read the other way: PrimeSim trains memory on
its own rollouts rather than on encoded ground-truth observations, so the model learns
the environment's response to motion it actually produced, which is what this suite
measures.

\subsubsection{EWMBench Evaluation}
\label{sec:ewmbench_exp}
EWMBench~\cite{yue2025ewmbench} evaluates embodied world models on eight
dimensions covering scene consistency (SceneC), motion accuracy (HSD, Dyn and
nDTW) and semantic alignment with the task instruction (BLEU, CLIP, Logics),
plus a Diversity term across candidates. The overall score is the sum of the
eight dimensions, each in $[0,1]$. We evaluate on the standard 21-episode
community split across 7 AgiBot World~\cite{agibot-world} tasks (as used
in~\cite{thinkerwm}). For each episode we generate multiple rollouts under a
fixed action-conditioned protocol, exploring a small family of sampler
settings so that both fidelity and cross-sample diversity are represented;
scores reflect EWMBench metric suite on these generations.

  These episodes are in-domain: AgiBot World contributes to the training corpus, and
  the models compared here are fine-tuned from the midtrained checkpoint on these
  datasets before scoring. The EWMBench numbers therefore measure how well the
  architecture absorbs an in-domain benchmark rather than zero-shot transfer to it;
  Section~\ref{sec:midtrain_analysis} is where held-out platforms are evaluated.

PrimeSim rolls out open-loop at 10\,FPS, conditioned on the first frame, the
task instruction and the ground-truth action trajectory, in 33-frame windows
chained over the episode. Videos are generated at $192\times256$, scored at
the EWMBench frame size of $480\times640$ on the head-camera stream against
the 10\,FPS ground-truth lane. These quality results use the multi-step
sampling recipe (50 denoising steps, CFG 1.5), while the wall-clock latency
numbers in Sec.~\ref{sec:inference_speed} use the four-step distilled
Stage-3 sampler.

\begin{table}[!ht]
  \centering
  \caption{EWMBench scores. Higher is better; Overall is the sum of the eight
  dimensions. The first block is action-conditioned like ours; the second
  block is instruction- or text-conditioned and therefore not iso-protocol
  with ours (e.g.\ Thinker-WM generates at 5\,FPS upsampled to 30\,FPS for
  scoring). GE-Sim rows are computed from the published Genie Envisioner
  simulation evaluation~\cite{genie-envisioner}, aligned to the standard
  EWMBench column names. Thinker-WM, Qwen-RobotWorld, LVP, Sora2 and Kling26
  rows are as reported on the same 21-episode split in~\cite{thinkerwm},
  compiled there from~\cite{qwen-robotworld}. $^\dagger$EnerVerse-FT is from
  the EWMBench paper's own evaluation~\cite{yue2025ewmbench}.
  $^\ddagger$Thinker-WM (action-cond.) conditions on the ground-truth action
  sequence; only its motion dimensions and Overall are
  reported~\cite{thinkerwm}, averaged over the evaluated checkpoints.}
  \label{tab:ewmbench}
  \setlength{\tabcolsep}{3pt}
  \small
  \begin{tabular}{l c c c c c c c c c}
    \toprule
    Method & SceneC $\uparrow$ & HSD $\uparrow$ & Dyn $\uparrow$ & nDTW $\uparrow$ & Div $\uparrow$ & BLEU $\uparrow$ & CLIP $\uparrow$ & Logics $\uparrow$ & Overall $\uparrow$ \\
    \midrule
    \multicolumn{10}{l}{\emph{Action-conditioned}} \\
    GE-Sim (LTX) \cite{genie-envisioner} & 0.900 & 0.940 & 0.780 & 0.980 & 0.011 & 0.330 & 0.908 & 0.970 & 5.82 \\
    GE-Sim (COSMOS2) \cite{genie-envisioner} & \textbf{0.910} & 0.870 & \textbf{0.850} & 0.970 & 0.010 & 0.310 & 0.902 & 0.970 & 5.79 \\
    Thinker-WM (action-cond.) \cite{thinkerwm}$^\ddagger$ & -- & 0.737 & 0.364 & 0.874 & -- & -- & -- & -- & 5.03 \\
    \textbf{PrimeSim (ours)} & 0.887 & \textbf{1.000} & 0.666 & \textbf{1.000} & 0.010 & \textbf{0.570} & \textbf{0.954} & \textbf{1.000} & \textbf{6.09} \\
    \midrule
    \multicolumn{10}{l}{\emph{Instruction- or text-conditioned (different protocol)}} \\
    EnerVerse-FT \cite{huang2026enerverse}$^\dagger$ & 0.943 & 0.536 & 0.536 & 0.596 & 0.069 & 0.180 & 0.864 & 0.978 & 4.70 \\
    Thinker-WM \cite{thinkerwm} & 0.888 & 0.564 & 0.363 & 0.710 & 0.012 & 0.269 & 0.902 & 0.952 & 4.66 \\
    Qwen-RobotWorld \cite{qwen-robotworld} & 0.914 & 0.566 & 0.343 & 0.671 & 0.011 & 0.208 & 0.883 & 1.000 & 4.60 \\
    LVP \cite{video-planner} & 0.880 & 0.425 & 0.043 & 0.623 & 0.009 & 0.218 & 0.900 & 0.952 & 4.05 \\
    Sora2 \cite{sora2} & 0.853 & 0.281 & 0.349 & 0.275 & 0.031 & 0.247 & 0.910 & 0.947 & 3.89 \\
    Kling26 \cite{kling} & 0.821 & 0.327 & 0.182 & 0.342 & 0.017 & 0.259 & 0.901 & 1.000 & 3.85 \\
    \bottomrule
  \end{tabular}
\end{table}

As shown in Table~\ref{tab:ewmbench}, PrimeSim reaches an overall score of
6.09, exceeding the action-conditioned GE-Sim (LTX, 5.82) reported under the
same metric accounting; the action-conditioned Thinker-WM variant stands at
5.03. The instruction- and
text-conditioned field sits further below: EnerVerse-FT reaches 4.70 in the
EWMBench paper's own evaluation, the strongest recent embodied world models
on the same 21-episode split, Thinker-WM (4.66) and Qwen-RobotWorld (4.60),
cluster just under it, and the strongest general video models remain below
3.9. We emphasize that the second block of the table is not
iso-protocol with ours, as those models condition on instructions or text
rather than actions.

\subsubsection{Inference Speed Comparison}
\label{sec:inference_speed}
We benchmark the wall-clock inference throughput of the four-step distilled
PrimeSim against recent action-conditioned world models, with CausalWM included
as a related text-conditioned generative reference.

\paragraph{Protocol.}
All timings are measured on a single NVIDIA A100-80GB in bfloat16 at batch size
one. The unit of work is one autoregressive chunk of future frames under each
model's released inference recipe, timed by wall clock with CUDA
synchronization, including GPU-to-CPU materialization when it is part of the
public API. Checkpoint loading, model construction and warm-up chunks are
excluded, and we report steady-state means; per-model warm-up schedules and
timing scopes are given in Appendix~\ref{sec:latency_appendix}. Because
multi-view generators emit several RGB images per time step, we report two
rates. \emph{Temporal FPS} (TempFPS) counts rollout time steps produced per
second. \emph{View-frame FPS} (ViewFPS) counts each camera view as one frame
and is our primary throughput metric: comparing only TempFPS would penalize a
multi-view model, since each time step already renders $V$ images. For
single-view models the two rates coincide; PrimeSim's ViewFPS is three times
its TempFPS.

\paragraph{Results.}
Table~\ref{tab:inference_speed} ranks the measured systems by ViewFPS. PrimeSim is
fastest: 30.35 ViewFPS (10.12 TempFPS), rendering a 33-frame three-view window at
$288\times384$ in 3.16\,s.

Among external systems the closest comparison is GE-Sim 2.0, which is also
four-step distilled and three-view. It reaches 6.73 ViewFPS, $1.67\times$ the
resolution- and horizon-matched BWM setting at 4.04 ViewFPS. Two published numbers
are not comparable on this bench and we do not use them: GE-Sim 2.0 reports
10.9\,Hz \cite{geniesim2} and DreamDojo 10.81\,FPS \cite{dreamdojo}, both measured
on H100 rather than A100, and DreamDojo's distilled student weights were not public
when we ran. BWM's own configuration ($672\times896$, 48 frames) falls to
0.78\,FPS, the clearest illustration of how strongly throughput depends on
resolution and horizon.

\paragraph{Caveats.}
The rows are not matched in resolution, horizon or denoising budget: PrimeSim
and GE-Sim 2.0 use four-step distilled samplers, while the public BWM and
Ctrl-World checkpoints run 50-step sampling. Few-step distillation is part of
the released systems, so the comparison is between released inference recipes
rather than matched compute. Ctrl-World generates at the lower native
$192\times320$ per view, and CausalWM is text- and first-frame-conditioned
rather than action-conditioned. PrimeSim's throughput also depends on rollout
length: longer continuous-KV episodes measure 7.63 TempFPS as the KV cache
grows, and we quote the short-window steady state. Finally, PrimeSim is timed
at $288\times384$ per view under the RoboTwin~2.0~\cite{robotwin2} Stage-3 evaluation contract,
whereas the quality results above use $192\times256$ generations; the latency
geometry is stated explicitly and is not shared with those tables. This
benchmark covers world-model chunk generation only, not an end-to-end control
loop with a policy.

\begin{table}[H]
  \centering
  \caption{Inference throughput on a single NVIDIA A100-80GB (bfloat16),
  measured as steady-state wall-clock throughput per autoregressive chunk under
  each method's released inference recipe. TempFPS counts rollout time steps
  per second; ViewFPS counts each camera view as one frame and coincides with
  TempFPS for single-view models. Rows are not matched in resolution, horizon
  or denoising steps. PrimeSim is measured at $288\times384$ per view under the
  RoboTwin~2.0 Stage-3 contract (33-frame windows, continuous KV, 32 usable
  frames per window), a different geometry from the $192\times256$ setting of
  the quality evaluations above. $^\dagger$CausalWM is text- and
  first-frame-conditioned, not action-conditioned.}
  \label{tab:inference_speed}
  \setlength{\tabcolsep}{2.5pt}
  \small
  \begin{tabular}{l c c c c c c}
    \toprule
    Model & Views & Res.\,/ view & Future $T$ & Steps & TempFPS $\uparrow$ & ViewFPS $\uparrow$ \\
    \midrule
    \textbf{Ours (PrimeSim Stage-3)} & 3 & $288\times384$ & 32 (33) & 4 & \textbf{10.12} & \textbf{30.35} \\
    GE-Sim 2.0 \cite{geniesim2}      & 3 & $384\times512$ & 25 & 4 & 2.24 & 6.73 \\
    BWM \cite{bwm} (fair-aligned)    & 1 & $384\times512$ & 25 & 50 & 4.04 & 4.04 \\
    Ctrl-World \cite{ctrl-world}     & 3 & $192\times320$ & 5 & 50 & 0.87 & 2.60 \\
    BWM \cite{bwm} (official)        & 1 & $672\times896$ & 48 & 50 & 0.75 & 0.75 \\
    CausalWM \cite{cwm}$^\dagger$    & 1 & $480\times640$ & 25 & 4+4+4 & 0.66 & 0.66 \\
    \bottomrule
  \end{tabular}
\end{table}

\subsection{What Midtraining Learns}
\label{sec:midtrain_analysis}

Midtraining is where the model acquires its action interface: action
conditioning and multiview coupling are introduced and trained over 20 dataset
families and 22 robot platforms (Sec.~\ref{sec:midtrain}). The question we ask
here is what else that stage leaves behind. If cross-embodiment training yields
transferable features rather than a fit to the training distribution, those
features should survive transfer along three axes at once: to platforms the model
has never driven, to scenes it has never seen, and to camera placements it has
never been mounted at. All three survive, and the first does so at a cost low
enough to change what adding a new robot means in practice: an action-following
rollout within 100 to 200 gradient steps, where the pretrained backbone does not
reach one at ten to twenty times the budget. At a learning rate of
    $3\times10^{-5}$, two hundred steps move the parameters by roughly $10^{-2}$, so
    this is a small amount of optimisation rather than a finetune to convergence. The
    comparison therefore tests what midtraining already provides, not how much can be
    recovered by fitting each new platform. The protocol is in
    Table~\ref{tab:few_shot_protocol}. Posttraining is analysed separately in
Sec.~\ref{sec:posttrain_analysis}, because it changes how the model is run
rather than what it has learned.

\begin{table}[H]
  \centering
  \caption{Few-shot adaptation to held-out embodiments. Both initialisations are
  finetuned under an identical recipe with single-trajectory batches at a batch
  size of 1 and a learning rate of $3\times10^{-5}$, and neither receives an
  embodiment identifier or any embodiment-specific parameters. Entries are gradient steps to an
  action-following rollout, meaning one in which the arm tracks the commanded
  trajectory and the scene responds to the motion actually realized.
  $^\dagger$No action-following rollout was reached within this budget.}
  \label{tab:few_shot_protocol}
  \begin{tabular}{l l c c}
    \toprule
    Platform & Morphology & Midtrained & Pretrained \\
    \midrule
    xArm 7      & 7-DoF single arm & 200          & 2000$^\dagger$ \\
    ViperX 300  & bimanual         & \textbf{100} & 2000$^\dagger$ \\
    UMI gripper & handheld         & 200          & 2000$^\dagger$ \\
    \bottomrule
  \end{tabular}
\end{table}

\begin{figure*}[!t]
  \centering

  \begin{subfigure}{\textwidth}
    \centering
    \includegraphics[width=\linewidth]{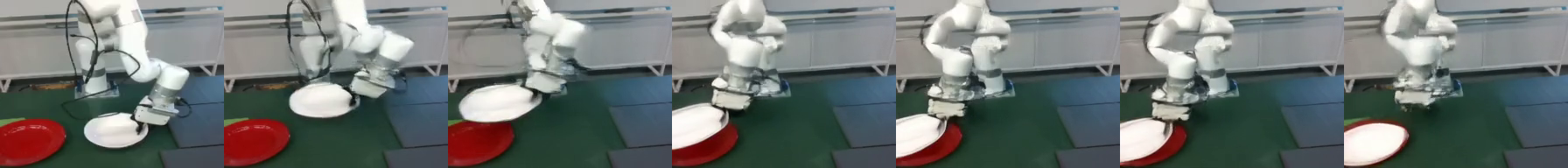}
    \caption{\emph{xArm 7}, midtrained, 200 gradient steps.}
    \label{fig:fewshot_xarm_mid}
  \end{subfigure}

  \vspace{2pt}

  \begin{subfigure}{\textwidth}
    \centering
    \includegraphics[width=\linewidth]{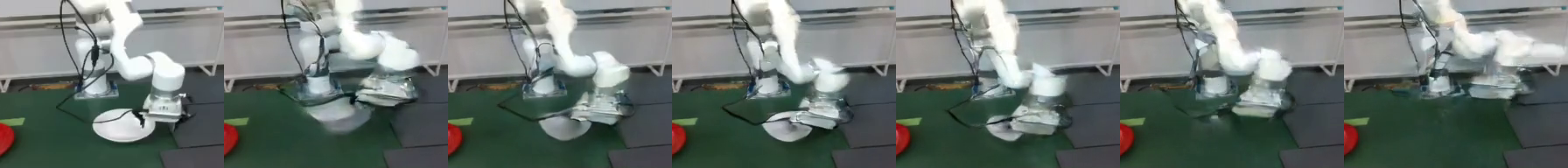}
    \caption{\emph{xArm 7}, pretrained backbone, 2000 gradient steps.}
    \label{fig:fewshot_xarm_pre}
  \end{subfigure}

  \vspace{2pt}

  \begin{subfigure}{\textwidth}
    \centering
    \includegraphics[width=\linewidth]{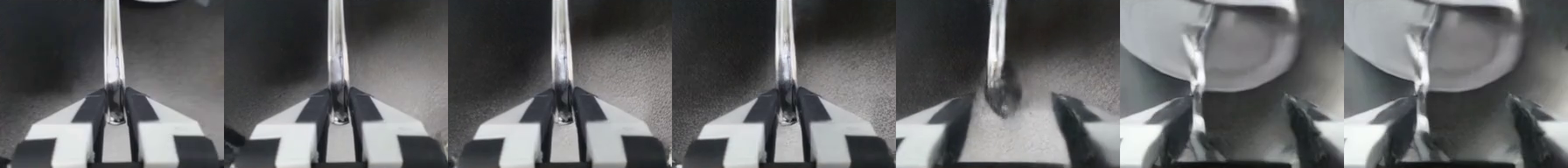}
    \caption{\emph{ViperX 300}, midtrained, 100 gradient steps.}
    \label{fig:fewshot_aloha_mid}
  \end{subfigure}

  \vspace{2pt}

  \begin{subfigure}{\textwidth}
    \centering
    \includegraphics[width=\linewidth]{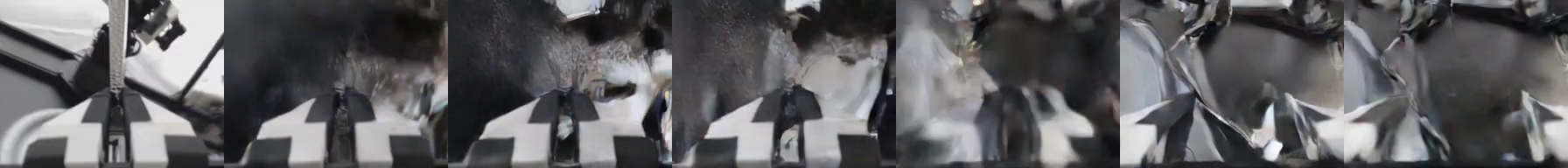}
    \caption{\emph{ViperX 300}, pretrained backbone, 2000 gradient steps.}
    \label{fig:fewshot_aloha_pre}
  \end{subfigure}

  \vspace{2pt}

  \begin{subfigure}{\textwidth}
    \centering
    \includegraphics[width=\linewidth]{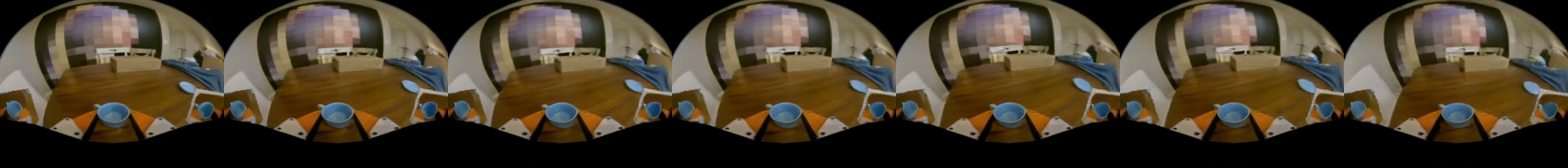}
    \caption{Original Stanford UMI gripper~\cite{umi}, midtrained, 200
    gradient steps.}
    \label{fig:fewshot_umi_mid}
  \end{subfigure}

  \vspace{2pt}

  \begin{subfigure}{\textwidth}
    \centering
    \includegraphics[width=\linewidth]{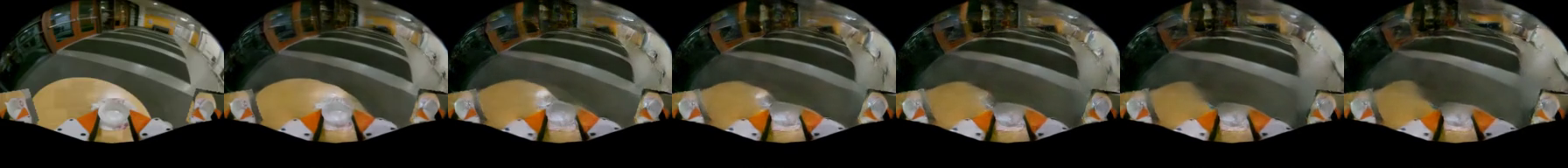}
    \caption{Original Stanford UMI gripper~\cite{umi}, pretrained backbone,
    2000 gradient steps.}
    \label{fig:fewshot_umi_pre}
  \end{subfigure}

  \captionsetup{font=footnotesize}
  \caption{Few-shot adaptation to unseen embodiments. Each pair shows the
  midtrained model above the pretrained backbone for one platform held out
  from training: a single 7-DoF arm (\emph{xArm 7}) drawn from Open
  X-Embodiment~\cite{oxe}, a bimanual platform (\emph{ViperX 300}, as used by
  ALOHA~\cite{aloha}) and the original Stanford UMI gripper~\cite{umi}. Similar
  embodiments do occur in the corpus, and they are what make
  transfer possible at all, since a wholly alien platform would leave nothing to
  build on. They are still new embodiments, however, and adaptation between
  embodiments does not happen for free: each brings its own kinematics, camera
  mounting and data distribution, which have to be acquired rather than
  inherited. All initialisations are finetuned
  under an identical recipe with single-trajectory batches at a batch size of 1,
  and none receives an embodiment identifier or any embodiment-specific
  parameters. The midtrained model reaches an action-following rollout within
  100 to 200 gradient steps on all three platforms. The pretrained backbone,
  which must initialise the action-conditioning and multiview modules from
  scratch, does not reach the same regime even at ten to twenty times that
  budget.}
  \label{fig:few_shot}
\end{figure*}

\paragraph{Embodiment Generalisation.}
Midtraining and posttraining are designed to broaden the model's coverage of
embodiments and dynamics. We probe whether that coverage has in fact yielded
transferable knowledge by asking what a new platform costs. We hold out three
platforms from the 20 dataset families of our training corpus
(Table~\ref{tab:datasets_frames}): a single 7-DoF arm (\emph{xArm 7}) drawn from
Open X-Embodiment~\cite{oxe}, a bimanual platform (\emph{ViperX 300}, as used by
ALOHA~\cite{aloha}) and the original Stanford UMI gripper~\cite{umi}. Similar
embodiments do appear in the corpus, through Franka in DROID~\cite{droid} and RoboMIND~\cite{robomind},
ALOHA-style simulation in RoboTwin~2.0, and the two UMI corpora, and that
similarity is what gives cross-embodiment training something to transfer. It does
not make adaptation automatic, however: the kinematics, camera mounting and data
distribution all differ, so whatever transfers must be re-expressed for the new
body, which is the gap a practitioner bridges when deploying to a new robot. UMI
is the sharpest case, since other handheld grippers appear in pre-training while
this device does not.

For each platform we finetune two initialisations under an identical recipe with
single-trajectory batches at a batch size of 1. Neither receives an embodiment
identifier or any embodiment-specific parameters, so the only difference between
them is the starting point: the midtrained model already carries the
action-conditioning and multiview modules learned from our 30{,}110-hour
cross-embodiment corpus (Sec.~\ref{sec:midtrain}), whereas the pretrained
Wan2.2-TI2V-5B backbone must initialise those modules from scratch. To avoid
penalising the baseline for this, we deliberately grant it ten to twenty times
the finetuning budget of the midtrained model: 2000 steps against 100 to 200.

The midtrained model reaches an action-following rollout on all three platforms
within 100 to 200 steps (Fig.~\ref{fig:few_shot}). The pretrained backbone, at
2000 steps, does not: its generations are neither visually plausible nor
action-controllable, remaining blurry and exhibiting no plausible movement or
contact.

We read this as transfer rather than fast fitting, for three reasons. The
optimiser is barely used: 200 gradient steps at a batch size of 1 draw 200
training windows from the recorded episodes, which is far too few to re-learn a
platform. No run receives an embodiment
identifier or an embodiment-specific parameter, so the gain cannot come from a
lookup on which body is present. And the same gap appears on a single arm, a
bimanual platform and a handheld device, which rules out an explanation tied to
one kinematic structure. The baseline, given ten to twenty times the optimisation,
never reaches the regime at all, so the difference is not one of budget. Taken
together, these leave little room for an account that does not involve knowledge
already carried in the weights. What they do not separate is corpus \emph{scale}
from the presence of an action interface at all: the baseline begins without one,
so the two cannot be disentangled here, and we make no claim about which
contributes more.

\paragraph{Scene and Camera-Pose Generalisation.}

The 30{,}110-hour corpus spans a wide range of environments, and we ask whether
that coverage transfers to a setting the model has never seen. We sample action
trajectories and initial frames from the XVLA open-source dataset \cite{xvla}, which is held
out of pre-training entirely. XVLA shares the AgileX Piper bimanual embodiment
and includes clothes-folding tasks, so the robot itself is familiar; what differs
is the camera configuration, the camera poses, and the setting as a whole. We
generate multiview video conditioned on the sampled actions and assess it along
three axes: \emph{action controllability}, whether the arms follow the commanded
trajectory; \emph{object-motion plausibility}, whether the cloth deforms and
travels as it should; and \emph{cross-view coherence}, whether the head and both
wrist views remain consistent with one another.

\begin{figure*}[t]
  \centering

  \begin{subfigure}{\textwidth}
    \centering
    \includegraphics[width=\linewidth]{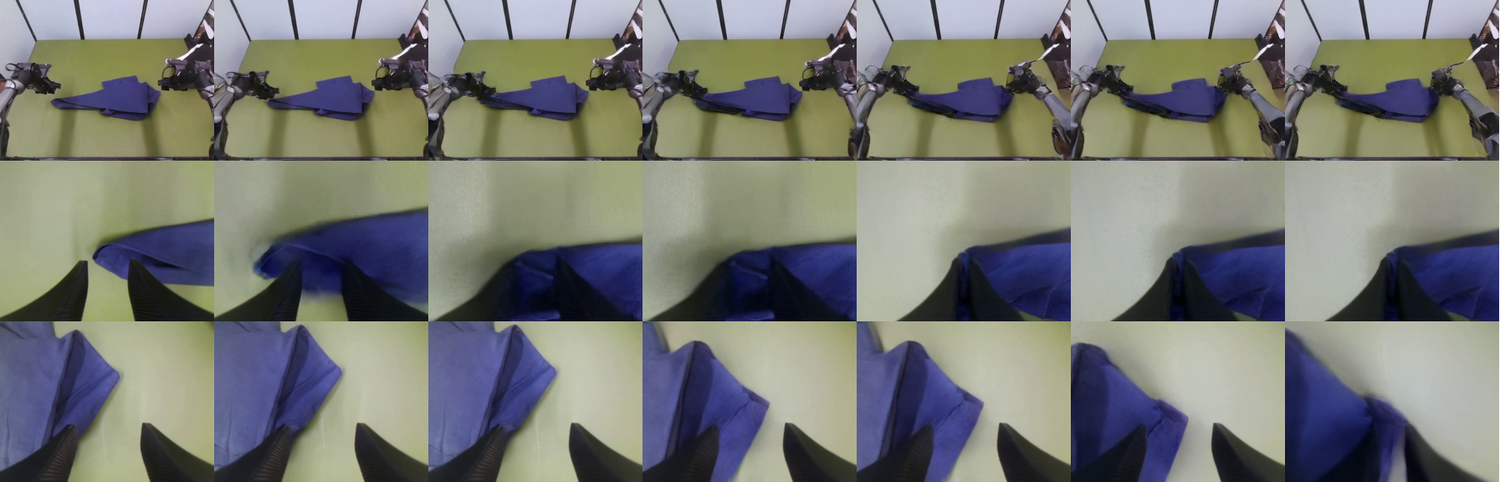}
    \caption{Before adaptation: zero-shot.}
    \label{fig:zero_shot_before}
  \end{subfigure}

  \vspace{4pt}

  \begin{subfigure}{\textwidth}
    \centering
    \includegraphics[width=\linewidth]{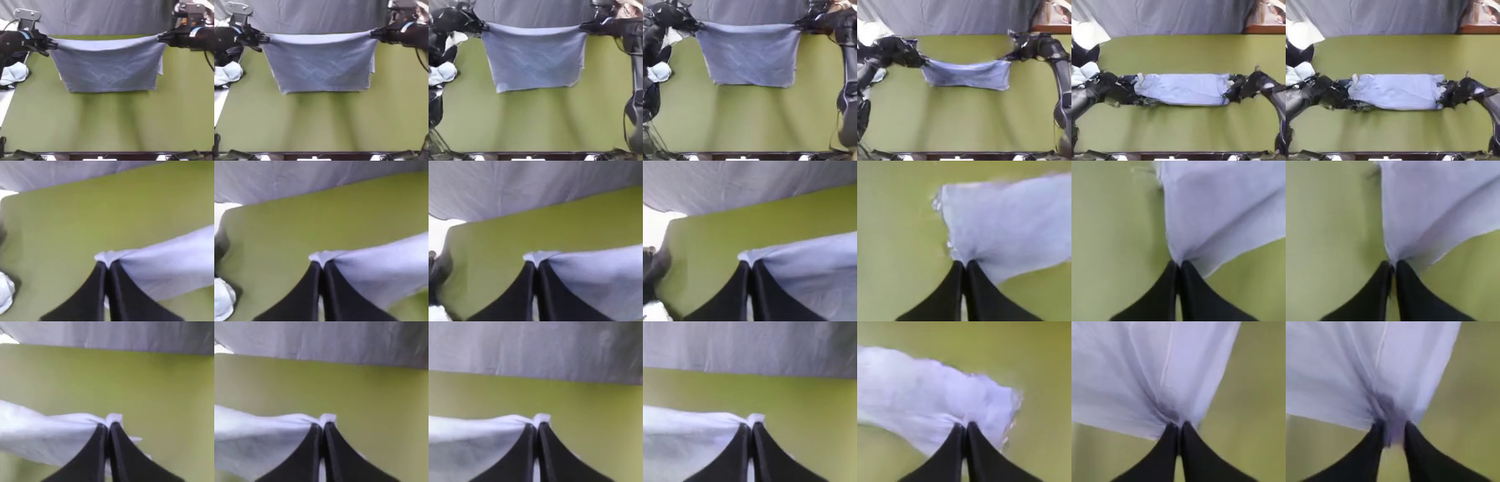}
    \caption{After 200 gradient steps at batch size 1.}
    \label{fig:finetune}
  \end{subfigure}

  \captionsetup{font=footnotesize}
  \caption{Zero-shot and few-shot generation on an unseen setting and camera
  pose. Both subfigures show the \emph{AgileX Piper} bimanual arm on XVLA
  \cite{xvla}, held out of pre-training entirely: the embodiment is in the
  corpus, but the scene, the garments and the camera placements are not. Rows are
  cameras (head, left wrist, right wrist) and columns are seven frames sampled
  uniformly from a single rollout. (a) Without adaptation the model controls the
  arms and renders the cloth, but the generated cloth state drifts from the
  commanded action once the motion becomes contact-rich. (b) After 200 gradient
  steps the arm trajectories and the cloth deformation track the commanded
  actions much more closely and the three views stay mutually consistent, so most
  of the cost of the transfer is recovered.}
  \label{fig:zero_shot}
\end{figure*}

Zero-shot, PrimeSim controls the arms accurately and generates simple cloth
motion faithfully, with quality degrading as the motion becomes contact-rich and
the generated cloth state drifts from the commanded action.
Table~\ref{tab:zero_shot} makes that cost explicit: against a matched
in-distribution baseline under an otherwise identical protocol, the model loses
3.97\,dB PSNR and 0.105 LPIPS on XVLA. Finetuning for 200 steps at a batch size of
1 recovers most of the gap, in both the arm trajectories and the cloth deformation
(Fig.~\ref{fig:zero_shot}). Seen platforms remain stronger than any transfer
setting, which is the expected ordering and is the point: these results measure
the cost of moving to a new platform, not a ceiling on what the simulator can
generate.

The two results together say something more specific than that the model
generalises. The same interface that carries to an unseen platform also carries
to an unseen setting, with the task prompt but no embodiment identifier at
inference. What transfers is therefore not a robot's appearance but the mapping
from actions to their visual consequences, which is what a simulator has to
encode and what a video prior alone does not.

\begin{table}[H]
  \centering
  \caption{Zero-shot transfer to a held-out dataset. Both rows share the
  checkpoint, sampler (50 steps, seed 42) and metrics at $192\times256$ over
  three jointly generated views; only the data differs. XR2~\cite{xr2} is held-out episodes
  of repositories seen in pre-training; XVLA is excluded and differs in scene,
  garments and camera placement. Means $\pm$ std over eight validation windows.
  SSIM is not comparable across the two datasets, so only PSNR and LPIPS should
  be read as a degradation.}
  \label{tab:zero_shot}
  \begin{tabular}{l l c c c}
    \toprule
    Data & Regime & PSNR $\uparrow$ & SSIM $\uparrow$ & LPIPS $\downarrow$ \\
    \midrule
    XR2  & in-distribution & $19.79 \pm 1.64$ & $0.731 \pm 0.053$ & $0.192 \pm 0.052$ \\
    XVLA & zero-shot       & $15.82 \pm 2.24$ & $0.762 \pm 0.049$ & $0.297 \pm 0.072$ \\
    \midrule
    \multicolumn{2}{l}{$\Delta$} & $-3.97$ & $+0.031$ & $+0.105$ \\
    \bottomrule
  \end{tabular}
\end{table}

\subsection{Posttraining Analysis}
\label{sec:posttrain_analysis}

Posttraining turns the midtrained model into a causal autoregressive simulator
through chunk-causal attention, few-step distillation and self-rollout training
(Sec.~\ref{sec:posttrain}). Posttraining primarily changes the model's causal 
generation and conditioning behavior, so we evaluate both its short-horizon
efficiency and fidelity, and its stability as the rollout horizon increases.

\subsubsection{Distillation Efficiency and Short-Horizon Fidelity}
\label{sec:distillation_exp}

Distillation serves two purposes in PrimeSim: it reduces inference latency for
interactive simulation, and it shortens the gradient graph during rollout
training, where backpropagating through 20 to 50 denoising steps does not fit on
current hardware. Table~\ref{tab:distillation_ablation} ablates the sampler on
the same RoboTwin~2.0 held-out split, with throughput measured on a single A100.
Relative to the 50-step causal Stage~1 teacher, the four-step Stage~3 model
improves throughput by approximately $9.0\times$; the measured rates for every
stage and for the released model are in the table.

\begin{figure*}[!t]
  \captionof{table}{Distillation results. Stage~1 is the causal autoregressive teacher at
  50 sampling steps. Stage~2 applies causal KV consistency distillation and
  Stage~3 causal KV distribution matching distillation (DMD), both reducing
  inference to 4 steps.}
  \label{tab:distillation_ablation}
  \small
  \setlength{\tabcolsep}{3pt}
  \makebox[\textwidth][c]{%
  \begin{tabular}{l c c c c c}
    \toprule
    Stage & Steps & FPS $\uparrow$ & PSNR $\uparrow$ &
    SSIM $\uparrow$ & LPIPS $\downarrow$ \\
    \midrule
    Stage 1 (Causal AR Diffusion) & 50 & 1.12 &
    \textbf{26.15} & \textbf{0.859} & \textbf{0.153} \\
    Stage 2 (CD Distillation) & 4 & 10.04 &
    24.40 & 0.841 & 0.193 \\
    Stage 3 (Asymmetric DMD) & 4 & \textbf{10.12} &
    24.87 & 0.838 & 0.176 \\
    \bottomrule
  \end{tabular}}
\end{figure*}

\subsubsection{Long-Horizon Robustness}
\paragraph{Episode-level adaptation.} To evaluate the final long-horizon adaptation stage, we directly compare
Stage~3, which performs self-rollout training within fixed-length clips, with
Stage~4, which extends training to complete robot episodes. Both models are
evaluated on the same ten episode-level held-out RoboTwin~2.0 trajectories using
identical initial observations and robot actions.

Each rollout is initialized from the first ground-truth RGB frame and
subsequently conditioned only on the model's own generated history. The
evaluated trajectories contain 610--1,046 generated frames, corresponding to
61.0--104.6 seconds at 10\,FPS, and include three synchronized camera views.

    Table~\ref{tab:long_horizon_robustness} reports both the average quality over
    the complete rollout and the degradation between the first and last complete
    32-frame generation windows. For PSNR and SSIM, degradation is computed as the
    first-window score minus the last-window score; for LPIPS, it is computed in the
    opposite direction. Lower degradation therefore indicates better long-horizon
    robustness for all three metrics.
    
    \begin{figure*}[!t]
  \captionof{table}{Long-horizon generation quality averaged over ten held-out RoboTwin~2.0
  episodes. Stage~3 uses fixed-length self-rollout training, while Stage~4
  extends self-rollout training to complete episodes. The Stage~4 result uses the rolling $(3+18)$ cache configuration.}
  \label{tab:long_horizon_robustness}
  \small
  \setlength{\tabcolsep}{3pt}
  \makebox[\textwidth][c]{%
  \begin{tabular}{lcccccc}
    \toprule
    Stage &
    Mean PSNR $\uparrow$ &
    Mean SSIM $\uparrow$ &
    Mean LPIPS $\downarrow$ &
    $\Delta$PSNR $\downarrow$ &
    $\Delta$SSIM $\downarrow$ &
    $\Delta$LPIPS $\downarrow$ \\
    \midrule
    Stage 3 & 12.572 & 0.401 & 0.654 &
    18.305 & 0.602 & 0.640 \\
    Stage 4 & \textbf{18.709} & \textbf{0.695} & \textbf{0.403} &
    \textbf{8.427} & \textbf{0.120} & \textbf{0.233} \\
    \bottomrule
  \end{tabular}}
\end{figure*}

\paragraph{Memory configuration ablation.}
We further compare rolling and hierarchical KV-cache organizations under the
same long-horizon evaluation protocol. For rolling memory, the allocation
denotes sink and rolling slots; for hierarchical memory, it denotes global,
slow, and local slots.

\begin{figure*}[!t]
  \captionof{table}{KV-cache organization ablation for long-horizon generation.}
  \label{tab:memory_configuration_ablation}
  \small
\setlength{\tabcolsep}{2pt}
\makebox[\textwidth][c]{%
\begin{tabular}{@{}lccccccccc@{}}
  \toprule
  & \multicolumn{3}{c}{Full-rollout average}
  & \multicolumn{3}{c}{Final window}
  & \multicolumn{3}{c}{Degradation} \\
  \cmidrule(lr){2-4}
  \cmidrule(lr){5-7}
  \cmidrule(lr){8-10}
  Memory configuration
  & PSNR $\uparrow$ & SSIM $\uparrow$ & LPIPS $\downarrow$
  & PSNR $\uparrow$ & SSIM $\uparrow$ & LPIPS $\downarrow$
  & $\Delta$PSNR $\downarrow$
  & $\Delta$SSIM $\downarrow$
  & $\Delta$LPIPS $\downarrow$ \\
  \midrule
  Rolling ($3+18$)
  & 18.709 & 0.695 & 0.403
  & 18.382 & 0.714 & 0.400
  & 8.427 & 0.120 & 0.233 \\
  Rolling ($0+21$)
  & 16.502 & 0.591 & 0.515
  & 12.973 & 0.459 & 0.670
  & 13.749 & 0.332 & 0.503 \\
  Rolling ($0+9$)
  & 15.220 & 0.547 & 0.577
  & 12.542 & 0.454 & 0.729
  & 13.641 & 0.353 & 0.548 \\
  Hierarchical ($3+9+9$)
  & \textbf{19.519} & \textbf{0.713} & \textbf{0.379}
  & \textbf{21.027} & \textbf{0.760} & \textbf{0.321}
  & \textbf{6.008} & \textbf{0.079} & \textbf{0.159} \\
  \bottomrule
\end{tabular}}
\end{figure*}

Under the same 21-slot budget, hierarchical memory consistently outperforms
the $3+18$ rolling cache. It improves both average and final-window quality,
while reducing $\Delta$PSNR from 8.427 to 6.008\,dB, $\Delta$SSIM from
0.120 to 0.079, and $\Delta$LPIPS from 0.233 to 0.159. This supports
separating persistent, long-term, and recent context within the KV cache.

\subsection{PrimeSim as Policy Simulator}
\label{sec:policy_simulator}

The video metrics in the preceding sections measure how closely PrimeSim
reproduces recorded trajectories. For a policy simulator, the more relevant
question is whether conclusions drawn inside it carry over to the real
environment. We examine this in two ways: whether PrimeSim predicts and ranks
the success rates of fixed policies as a physics simulator does
(Sec.~\ref{sec:policy_eval}), and whether it can serve as the training
environment for reinforcement-learning post-training of a policy
(Sec.~\ref{sec:policy_rl}).

\paragraph{Setup.}
We use the LIBERO-Spatial suite (10 tasks) in the LIBERO simulator as the
reference, where success is LIBERO's privileged-state task predicate. Policies
are OpenVLA-OFT after supervised finetuning on LIBERO-Spatial demonstrations;
they predict 8-step action chunks from the main camera view. Inside a world model, 
success is judged by a ResNet-based binary classifier finetuned on LIBERO-Spatial frames,
which predicts from each generated frame whether the task has been completed; 
an episode receives a sparse reward of 1 if any of its frames is classified as success
and 0 otherwise.

\subsubsection{Policy Evaluation Fidelity}
\label{sec:policy_eval}

We first freeze the policy and ask whether the world model predicts its
success rate. We take eight OpenVLA-OFT policies of clearly different
competence, whose simulator success ranges from 29.3\% to 85.0\%. Each policy
is run with greedy actions from the same 300 initial states (10 tasks $\times$
30 trials), once in the LIBERO simulator and once closed-loop inside the world
model. As a reference world model we use a publicly released
Wan2.2-TI2V-5B~\cite{wan} finetuned on LIBERO-Spatial
(Wan2.2-FT). We report the Spearman rank correlation $\rho$ and the Pearson
correlation $r$ between world-model and simulator success, and the mean
absolute gap $|\Delta|$ between the two.

\begin{figure}[!t]
  \centering
  \begin{minipage}[b]{0.44\linewidth}
    \centering
    \includegraphics[width=\linewidth]{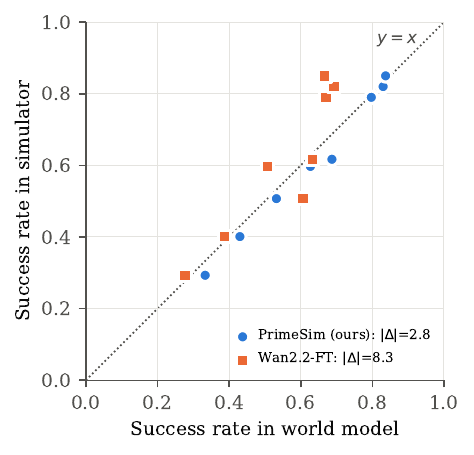}
    {\small (a) Success in world model vs.\ simulator}
  \end{minipage}\hfill
  \begin{minipage}[b]{0.44\linewidth}
    \centering
    \includegraphics[width=\linewidth]{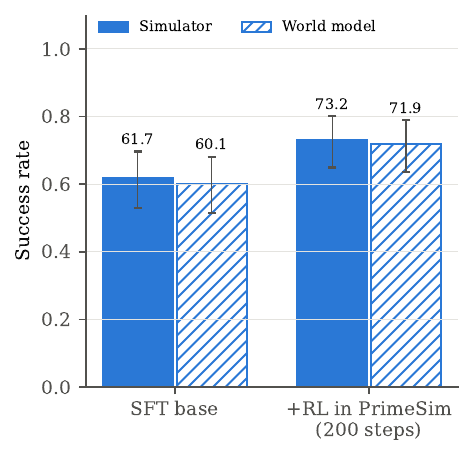}
    {\small (b) Before vs.\ after RL post-training in PrimeSim}
  \end{minipage}
  \caption{\textbf{(a)} Each point is one OpenVLA-OFT policy, evaluated on the
  same 300 LIBERO-Spatial initial states in the world model (x) and in the
  LIBERO simulator (y). The dotted line marks perfect agreement; points above
  it mean the world model under-estimates the policy. \textbf{(b)} Success rate
  of the SFT policy before and after 200 steps of GRPO post-training inside
  PrimeSim, measured in the simulator (solid) and in PrimeSim (hatched).}
  \label{fig:wm_rl_correlation}
\end{figure}

PrimeSim orders all eight policies exactly as the simulator does (Spearman
$\rho=1.00$, Pearson $r=0.99$) and stays within 7.0 points of the simulator
for every policy, with a mean absolute gap of 2.8 points
(Fig.~\ref{fig:wm_rl_correlation}(a)). Wan2.2-FT also tracks the overall trend
($r=0.91$) but mis-orders several policies ($\rho=0.90$), and its errors grow
with competence: it under-estimates the three strongest policies, whose
simulator success is 79.0\%, 82.0\% and 85.0\%, by 12.0, 12.7 and 18.3
points, for a mean absolute gap of 8.3 points. A world model that
systematically under-rates good policies is of limited use for checkpoint
selection, which is exactly where policy evaluation in a world model is needed.

\subsubsection{RL Post-training inside PrimeSim}
\label{sec:policy_rl}

Beyond evaluation, PrimeSim can serve directly as the training environment for
policy post-training. Starting from the SFT policy, we run GRPO (group size 8)
entirely inside PrimeSim: rollouts are generated closed-loop by the world model
and receive the sparse binary reward from the success classifier, and the 
physics simulator is never queried during training; it is used only to evaluate the resulting policy.
After 200 RL steps, the policy's success rate in the LIBERO simulator rises
from 61.7\% to 73.2\%, and PrimeSim's own estimate (60.1\% to 71.9\%) stays
within 1.6 points of it (Fig.~\ref{fig:wm_rl_correlation}(b)), showing that
improvements obtained purely inside PrimeSim transfer to the physics simulator.

\paragraph{Scope of this experiment.}
The aim here is to show that PrimeSim supports RL post-training, not to maximise
absolute success on LIBERO. We deliberately use a short training budget and an
off-the-shelf success classifier, so the numbers measure whether closed-loop policy
optimisation inside the model works at all, not what the best attainable score is.
What longer optimisation would buy, and what it would risk, is discussed under
limitations in Section~\ref{sec:conclusions}.

\subsection{Ablation Study}

We run component ablations rather than stage comparisons. Action representation,
action conditioning and multiview consistency each isolate one midtraining choice,
and we train all three from the pretrained checkpoint rather than the midtrained
one, so that the stronger priors midtraining already supplies cannot mask the
contribution of the component under test. The distillation ablation is the
exception, since post-training begins from the midtrained model by construction
and its variants are therefore compared from that checkpoint. The midtraining
ablations use a small in-house dataset collected for folding clothes on one
    humanoid embodiment.

This setup has a scope worth stating once rather than in every subsection. Each
ablation below is run on a single embodiment, an in-house humanoid for some and the
simulated RoboTwin~2.0~\cite{robotwin2} setup for others, so the comparisons isolate how one component
behaves within one kinematic frame; none of them measures cross-embodiment
behaviour, and the platform adaptation results in Sec.~\ref{sec:midtrain_analysis}
are the evidence that speaks to that. A mechanism whose value is specifically
cross-embodiment, such as the relative action convention cancelling each robot's
absolute origin, has nothing to act on in either setting, so these numbers are
more likely to understate such components than to overstate them.

\subsubsection{Action Representation}
We compare three ways of representing actions: absolute poses, delta poses
relative to the previous frame, and our relative action chunk anchored to the
first frame of the window. To isolate the effect of the representation itself,
we initialize all variants from the pretrained checkpoint rather than the
midtrained one, so the comparison is not biased by stronger priors learned
during midtraining. Results are shown in Table~\ref{tab:relative_action}, reported in PSNR~\cite{psnr}, SSIM~\cite{ssim}, and
LPIPS~\cite{lpips}.

\begin{table}[H]
  \centering
  \caption{Ablation study on action representation. Best result per metric in
  \textbf{bold}. $\uparrow$: higher is better; $\downarrow$: lower is better.}
  \label{tab:relative_action}
    \begin{tabular}{l c c c}
      \toprule
      Action Representation & PSNR $\uparrow$ & SSIM $\uparrow$ & LPIPS $\downarrow$ \\
      \midrule
      Absolute action       & 22.23 & 0.79 & 0.14 \\
      Delta action          & 21.34 & 0.81 & 0.18 \\
      Relative action chunk & \textbf{24.56} & \textbf{0.84} & \textbf{0.13} \\
      \bottomrule
    \end{tabular}
\end{table}

\subsubsection{Action Conditioning}
\label{sec:action_cond_exp}

\begin{figure*}[t]
  \centering
  \begin{subfigure}{\textwidth}
    \centering
    \includegraphics[width=\linewidth]{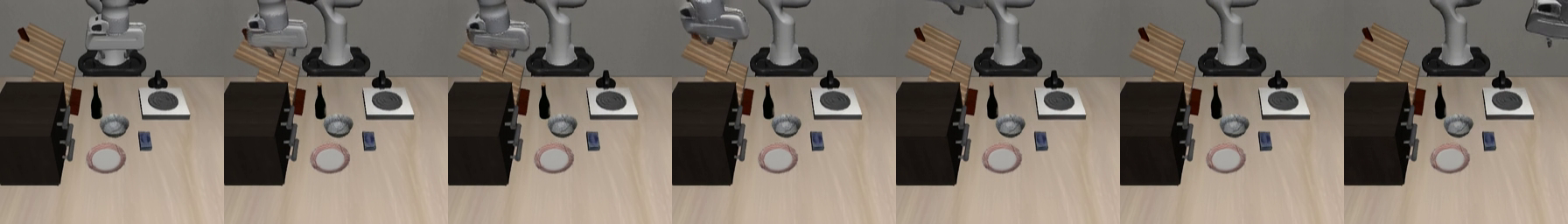}
    \caption{}
    \label{fig:cf_a}
  \end{subfigure}

  \vspace{2pt}

  \begin{subfigure}{\textwidth}
    \centering
    \includegraphics[width=\linewidth]{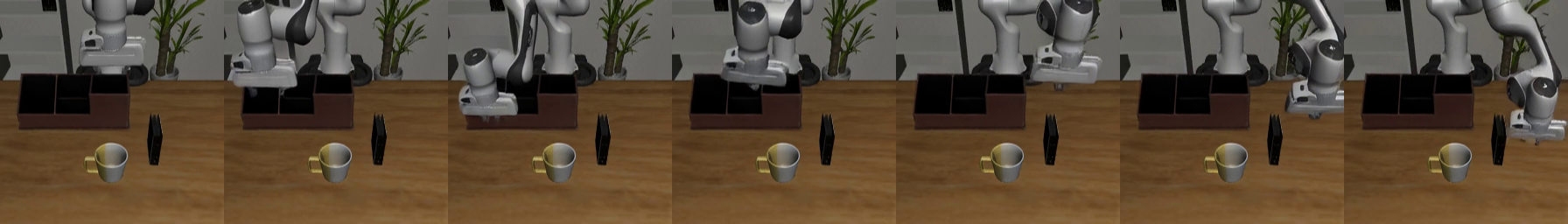}
    \caption{}
    \label{fig:cf_b}
  \end{subfigure}

  \caption{Counterfactual action following on LIBERO. Each row shows seven
  frames sampled uniformly from a single PrimeSim rollout, generated
  conditional on an action sequence rather than a task prompt. Our evaluation
  is built on counterfactual commands, meaning action trajectories that do not
  correspond to the demonstrated behaviour. Faithful generation therefore
  requires the model to respond to the commanded motion rather than replay the
  recorded one. Static scene elements remain stable across the rollout while
  the commanded motion is carried out, indicating that the generated future is
  driven by the action rather than by visual plausibility alone.}
  \label{fig:counterfactual}
\end{figure*}

We ablate our two action conditioning mechanisms, framewise action
conditioning and the frame-action adapter, by applying each in isolation and
then together. Because prior work~\cite{bwm, wovr} also uses two-channel action
conditioning but relies on standard cross-attention and AdaLN-zero~\cite{dit},
we compare each proposed channel against its conventional counterpart:
framewise conditioning against vanilla cross-attention, and the token-level
adapter against AdaLN. The final row tests both proposed channels together.
Results are shown in Table~\ref{tab:action_cond_ablation}.

Reconstruction metrics do not measure controllability, so we test action
following directly. We evaluate on counterfactual actions, meaning action
sequences that do not correspond to the recorded trajectory: faithful generation
therefore requires the model to respond to the commanded motion rather than
replay the recorded one. PrimeSim follows these out-of-distribution commands, as
shown in Fig.~\ref{fig:counterfactual}, where static scene elements remain
stable across the rollout while the commanded motion is carried out.

A qualitative figure cannot by itself settle whether the motion is genuinely
action-driven, and we do not ask it to. WorldSimProbe provides the independent
quantitative test, and its T3 suite, action-source behaviour preservation, asks
precisely the counterfactual question: whether behaviour survives when the
actions originate from a different source than the one that produced the video.
Against five open-source action-conditioned world models, PrimeSim ranks first on
T3 (72.3) and on the coarse trajectory axis T2 (67.9), and second on local
calibration T1 (70.2 against LingBot-VA's 77.0)
(Table~\ref{tab:worldsimprobe}). The two lines of evidence agree, and they agree
on the residual weakness as well: action following holds at trajectory scale,
while fine-grained local calibration is not yet the best in the field.

\begin{table}[H]
  \centering
    \caption{Ablation study on action conditioning. The \textit{Baseline} column
    uses vanilla cross-attention for framewise conditioning and Adaptive LayerNorm
    (AdaLN) for the adapter; \textit{Ours} uses modified cross-attention and
    token fusion and shifting, respectively. The final row combines both
    mechanisms. Higher PSNR is better.}
  \label{tab:action_cond_ablation}
    \begin{tabular}{l c c c}
      \toprule
      \multicolumn{2}{c}{Action Conditioning Mechanism} & \multicolumn{2}{c}{PSNR $\uparrow$} \\
      \cmidrule(lr){1-2} \cmidrule(lr){3-4}
      Framewise & Adapter & Baseline & Ours \\
      \midrule
      \checkmark & $\times$   & 23.23 & 24.10 \\
      $\times$   & \checkmark & 20.10 & 20.24 \\
      \checkmark & \checkmark & 23.31 & 24.56 \\
      \bottomrule
    \end{tabular}
\end{table}

\clearpage

\subsubsection{Multiview Consistency}

\begin{figure*}[t]
  \centering
  \includegraphics[width=\textwidth]{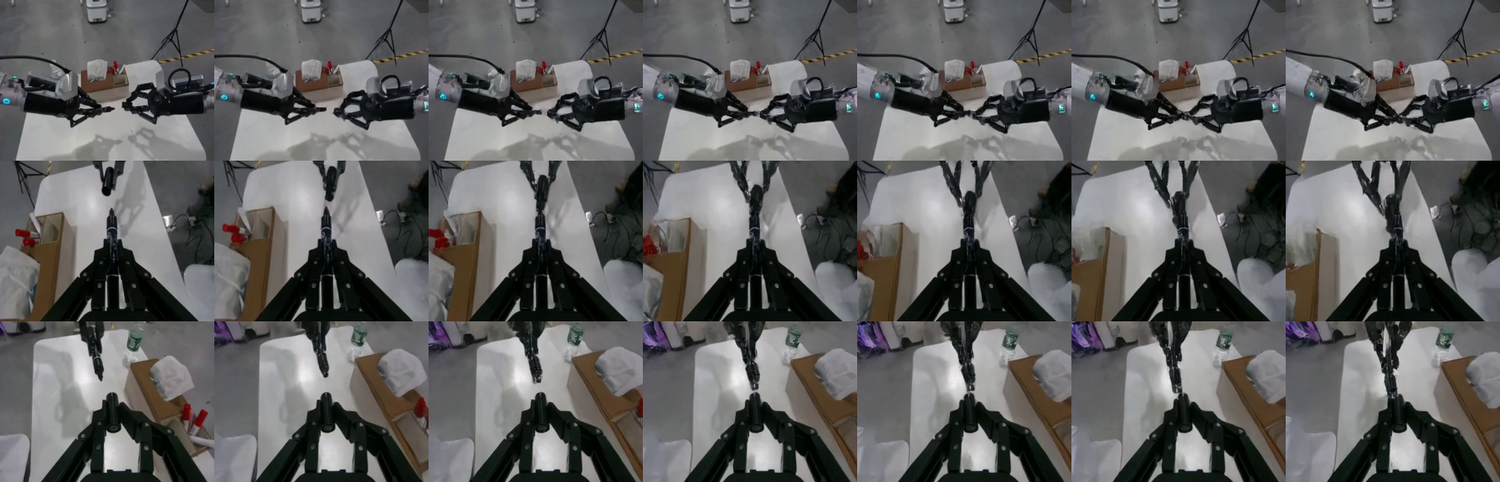}
  \caption{Joint multiview generation. Rows are cameras, from top to bottom
  head, left wrist and right wrist; columns are seven frames sampled uniformly
  across a single 33-frame PrimeSim rollout. The three views are generated
  together under a single action sequence rather than independently, and all
  respond to the same commanded motion. The wrist views change most over the
  interval, consistent with their proximity to the manipulation, while the head
  view tracks the arms throughout.}
  \label{fig:multiview}
\end{figure*}
The three views share one transformer stack rather than separate per-camera
networks. Each of the 30 DiT blocks attends within its own camera view, and every
third block instead spans all three, so relations inside a view are established
before information is periodically integrated across views. A learned camera token
marks which view each token belongs to.

We perform an incremental ablation of the multiview consistency mechanisms,
introducing each component in turn to isolate its contribution. The baseline
concatenates frames along the height dimension without any conditioning. We
then add a view token that conditions each view, and finally apply intra-view
attention across the frame tokens. Results are shown in Table~\ref{tab:multiview_ablation}.

\begin{table}[H]
  \centering
    \caption{Ablation study on multiview consistency. Components are added
    incrementally from naive frame concatenation. Best result in \textbf{bold}.}
  \label{tab:multiview_ablation}
    \begin{tabular}{l c c c}
      \toprule
      Multiview Mechanism & PSNR $\uparrow$ & SSIM $\uparrow$ & LPIPS $\downarrow$ \\
      \midrule
      Frame concatenation only & 16.30 & 0.62 & 0.58 \\
      + View token conditioning & 23.78 & 0.83 & 0.18 \\
      + Intra-view attention & \textbf{24.56} & \textbf{0.84} & \textbf{0.15} \\
      \bottomrule
    \end{tabular}
\end{table}

\clearpage

\subsection{Qualitative Results}
We show rollouts that are easier to judge by eye than by metric, and that the
quantitative sections do not otherwise cover.

\begin{figure}[H]
  \centering
  \includegraphics[width=\linewidth]{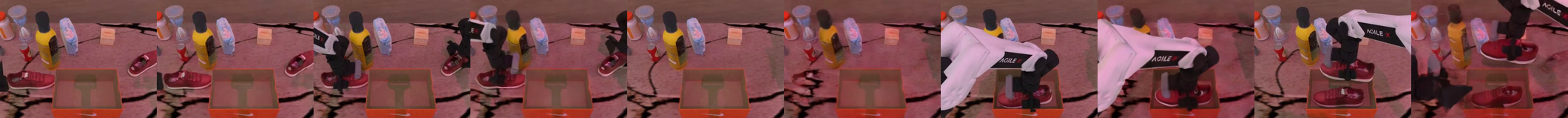}

  \par\vspace{2pt}
  \captionsetup{font=footnotesize}
  \caption{\emph{Memory under occlusion.} Five frames from a single
  RoboTwin~2.0 rollout, head camera. The arm sweeps across the workspace and its hand
  passes in front of the shoe in the lower left (fourth frame), so the shoe leaves
  the camera's view entirely. When the hand clears, the shoe is present again at
  the same position, scale and orientation rather than being re-invented or
  dropped. Each window is conditioned on a selected KV-cache memory rather than on
  the previous frame alone, so an object occluded for several seconds is carried
  forward instead of being lost when it re-enters view.}
  \label{fig:robotwin_memory}
\end{figure}

First, a sequence isolates what the temporal memory buys under occlusion
(Fig.~\ref{fig:robotwin_memory}).

\begin{figure}[H]
  \centering
  \includegraphics[width=\linewidth]{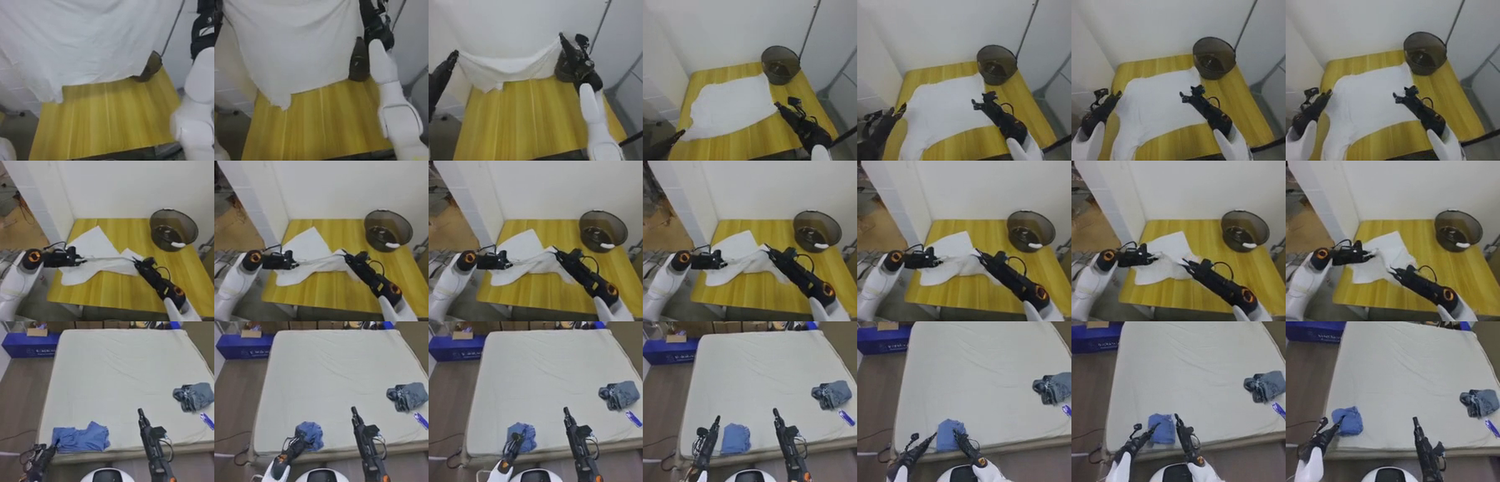}

  \par\vspace{2pt}
  \captionsetup{font=footnotesize}
  \caption{Cloth manipulation rollouts, seven frames sampled uniformly
  from each. All three are AgiBot G2 humanoid episodes shown from the head camera:
  a white cloth on a bamboo table beside a metal bowl (top two rows), and a blue
  cloth on a bedding surface with a folded garment to its right (bottom row). The
  garment deforms continuously across each rollout and stays distinguishable from
  the surface beneath it rather than dissolving into the background, which is the
  failure mode deformable objects expose over long horizons.}
  \label{fig:cloth_rollouts}
\end{figure}

A second sequence shows rollouts on deformable cloth, the material class where
long-horizon coherence is hardest to maintain (Fig.~\ref{fig:cloth_rollouts}).

\begin{figure}[H]
  \centering
  \includegraphics[width=\linewidth]{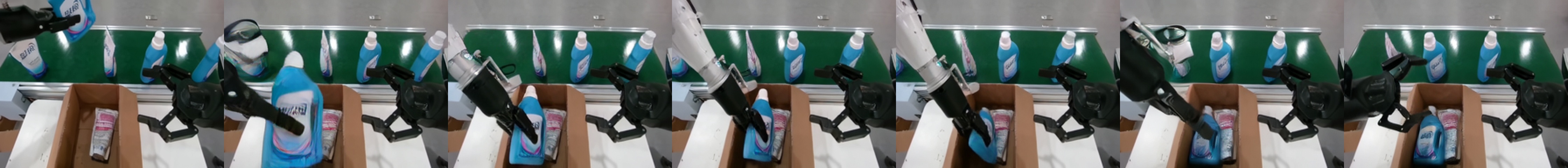}

  \vspace{2pt}

  \includegraphics[width=\linewidth]{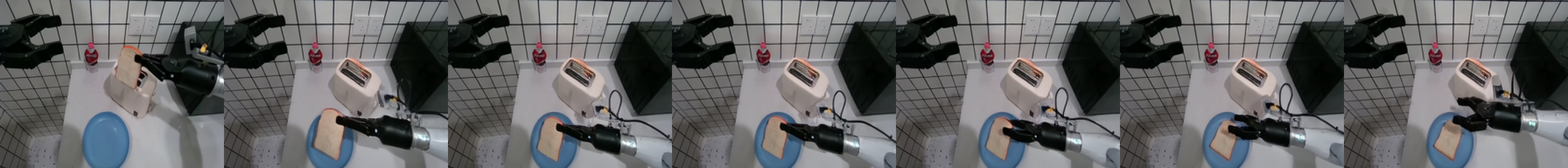}

  \par\vspace{2pt}
  \captionsetup{font=footnotesize}
  \caption{Generated rollouts on two EWMBench tasks, the qualitative companion
  of Table~\ref{tab:ewmbench} (Sec.~\ref{sec:ewmbench_exp}). Each row shows
  seven frames sampled uniformly from a single rollout, generated open-loop at
  10\,FPS from the first frame, the task instruction and the ground-truth
  action trajectory, and shown at the $480\times640$ frame size used for
  scoring. Top: the arm picks a bottle from the conveyor and packs it into a
  cardboard box already holding other items. Bottom: the arm places a slice of
  bread onto the plate beside the toaster.}
  \label{fig:ewmbench_rollouts}
\end{figure}

Two further sequences give a qualitative view of the EWMBench generations behind
Table~\ref{tab:ewmbench} (Fig.~\ref{fig:ewmbench_rollouts}).

\FloatBarrier

\clearpage

\section{Related Work}
\label{sec:related_works}
\paragraph{Video Generation in Robotics}
Recent advances in general-purpose video generation \cite{wan, wow, cogvideox} have catalyzed a growing body of work applying these models to robotics. One prominent direction leverages video generation for data expansion: by synthesizing diverse visual trajectories, these models augment limited teleoperation data, enabling broader coverage of environments, objects, and tasks \cite{cosmos, real2edit2real, qwen-robotworld, xiaomi}. This mitigates a central bottleneck in robot learning—the scarcity of large-scale, diverse demonstrations. A complementary direction exploits the spatiotemporal physical understanding that video models acquire through large-scale pretraining. Rather than treating generated video merely as data, these works use the model's internal representations as a backbone for action prediction and planning, demonstrating strong performance on downstream robotic tasks \cite{cosmos3, genie-envisioner, uva, motus, video-planner, dyna2, cosmos-policy, mimic-video, dreamzero}. Together, these two lines of work reflect a broader shift: video generation models are evolving from passive data synthesizers into active components of the robot learning pipeline, serving simultaneously as sources of supervision and as engines of physical reasoning.

\paragraph{Interactive World Models}
Closely related to video generation is the emerging area of interactive world models, in which generation is conditioned on and steered by actions \cite{lingbot-world, lingbot-world2, diamond, gamengen, matrix, yume-15, genie, hunyuan-gamegraft, echo-joy}. Much of this work targets interactive video games, from early action-conditioned systems \cite{diamond, genie} to more recent higher-fidelity, longer-horizon ones \cite{matrix, gamengen, hunyuan-gamegraft, yume-15, echo-joy}. More recently, the paradigm has extended to robotics \cite{geniesim2, rise, ctrl-world, world-action-verifier, worldeval, dworldeval, interactive-world-simulator, world-vla-loop, wovr, vlaw, gigabrain-05m, gemini-wm-eval, sc3-eval, dreamdojo}. These robot-centric models differ from their game-oriented counterparts along several axes: they emphasize multi-view consistency \cite{geniesim2, rise, ctrl-world} rather than a single fixed viewpoint; they operate on continuous, physically grounded control signals \cite{ctrl-world, world-action-verifier} rather than discrete symbolic inputs \cite{diamond, genie}; and they prioritize geometric and physical consistency \cite{worldeval, dworldeval, interactive-world-simulator} over visual aesthetics and responsiveness \cite{matrix, hunyuan-gamegraft}. Their use cases also diverge: game-based models target entertainment and content creation, whereas robot-based ones serve as learned simulators providing synthetic feedback for action evaluation, policy learning, and planning \cite{world-vla-loop, wovr, vlaw, gigabrain-05m, gemini-wm-eval, sc3-eval, dreamdojo}.

\section{Limitations and Conclusions}
\label{sec:conclusions}

\paragraph{Limitations.}
Our evaluation is broad but not exhaustive, and we prefer to state the gaps rather
than let them be inferred. One concerns the model itself: PrimeSim generates at
$192\times256$ per view, so our claims concern simulator fidelity, whether actions
are realized and the environment responds, rather than pixel-level realism, and we
make no like-for-like visual fidelity comparison against models operating at higher
resolution.

A further gap is a question of resources rather than of method. Adaptation to a new
platform is few-shot, requiring 100 to 200 gradient steps on that platform's own
data, rather than zero-shot. We expect this gap to close with more midtraining
GPU-hours, tighter alignment of each corpus's action and state conventions, and a
larger and more diverse corpus, since all three act on the same quantity that
makes the interface transferable at all. We have not demonstrated it, and state it
as an expectation rather than a result.

A final gap concerns scope, and it is the one we consider most important.
Most of our quantitative evidence, and all of our closed-loop evidence, comes
from simulation: the benchmark comparisons are on LIBERO, and policy training
and evaluation inside PrimeSim are validated in simulation only. Our
real-platform results establish that PrimeSim generates faithful video on real
data, zero-shot on XVLA and after few-shot adaptation to xArm~7, ViperX~300 and
a UMI gripper, and that it follows commanded action across the whole action
space including chassis velocity. What they do not yet establish is the
simulator role on physical hardware. Whether generated rollouts are faithful
enough to stand in for simulator rollouts when improving a policy on a real
platform remains open, and it is the direction we intend to pursue next.

\paragraph{Conclusions.}
We presented PrimeSim, a world model that turns a video generation backbone into
an action-conditioned, multiview, causal simulator of robot manipulation. A
three-stage curriculum does the work: midtraining on a 30{,}110-hour
cross-embodiment corpus, midtraining that introduces action and multiview
conditioning, and posttraining that converts the model into a causal few-step
autoregressive generator. The result conditions on actions and the task
prompt, but on no embodiment identifier at inference. Against five open-source
action-conditioned world models, PrimeSim leads the WorldSimProbe axes that
measure action realization at trajectory scale and interaction dynamics, and it
reaches those results at 10.1 FPS through four-step distillation rather than the
fifty steps the teacher requires. Its features transfer: three held-out
platforms reach action-following rollouts within 100 to 200 gradient steps,
where a general video prior given ten to twenty times that budget does not. We
expect the combination of a large cross-embodiment corpus, an explicit action
interface and a causal few-step sampler to be a practical route toward learned
simulators that are fast and controllable enough to train policies inside.

\clearpage

\bibliographystyle{plainnat}
\bibliography{reference}

\appendix

\subsection{Corpus Composition}

\newcommand{\na}{--}

\begin{table*}[t]
\centering
\small
\setlength{\tabcolsep}{5pt}
\caption{%
  Training data mixture, measured from dataset metadata rather than from
  upstream release figures. Hours are derived as
  $\text{frames}/\text{fps}/3600$ and so are consistent with the frame counts
  by construction. The mixture spans 20 dataset families, 1{,}773 LeRobot
  repositories, 1.52M episodes and 22 distinct robot platforms; it is
  84\% real by trajectory time. \emph{Real} and \emph{Sim} split
  each family by data source, and two families contribute both.
}
\label{tab:datasets_frames}
\begin{tabular}{@{}lrrrrrr@{}}
\toprule
& \multicolumn{3}{c}{Hours} & \multicolumn{3}{c}{Frames} \\
\cmidrule(lr){2-4}\cmidrule(lr){5-7}
Dataset family & Real & Sim & Total & Real & Sim & Total \\
\midrule
In-House Data                          & 9,540 & \na   & 9,540 & 375M & \na    & 375M \\
InternData-A1 \cite{interndata-a1}     & 4,000 & 170   & 4,170 & 410M & 1M     & 411M \\
AgiBot World \cite{agibot-world}       & 2,300 & \na   & 2,300 & 243M & \na    & 243M \\
Hy-Embodied (UMI) \cite{hy-vla}        & 2,100 & \na   & 2,100 & 226M & \na    & 226M \\
BEHAVIOR-1K 2026 \cite{behavior1k}     & \na   & 2,000 & 2,000 & \na  & 211M & 211M \\
HiFi-UMI-130k \cite{hifi-umi}          & 2,000 & \na   & 2,000 & 180M & \na    & 180M \\
InternData-M1 \cite{interndata-m1}     & 160   & 1,700 & 1,860 & 18M  & 181M & 199M \\
XR2 \cite{xr2}                         & 1,500 & \na   & 1,500 & 162M & \na    & 162M \\
RoboMIND \cite{robomind}               & 850   & \na   & 850   & 92M  & \na    & 92M  \\
RoboCOIN \cite{robocoin}               & 700   & \na   & 700   & 77M  & \na    & 77M  \\
AgiBot World 2026 \cite{agibot2026}    & 610   & \na   & 610   & 66M  & \na    & 66M  \\
RoboCasa \cite{robocasa}               & \na   & 610   & 610   & \na  & 44M    & 44M  \\
DROID \cite{droid}                     & 510   & \na   & 510   & 28M  & \na    & 28M  \\
Galaxea OpenWorld \cite{galaxea}       & 490   & \na   & 490   & 26M  & \na    & 26M  \\
MimicGen \cite{mimicgen}               & \na   & 210   & 210   & \na  & 15M    & 15M  \\
Kai0 \cite{kai0}                       & 130   & \na   & 130   & 15M  & \na    & 15M  \\
RoboTwin~2.0~\cite{robotwin2}          & \na   & 60    & 60    & \na  & 6M     & 6M   \\
RT-1 Fractal \cite{rt-1}               & 350   & \na   & 350   & 4M   & \na    & 4M   \\
Bridge V2 \cite{bridgev2}              & 110   & \na   & 110   & 2M   & \na    & 2M   \\
LIBERO \cite{libero}                   & \na   & 10    & 10    & \na  & 1M     & 1M   \\
\midrule
\textbf{Total} & \textbf{25,350} & \textbf{4,760} & \textbf{30,110} & \textbf{1,924M} & \textbf{459M} & \textbf{2,383M} \\
\bottomrule
\end{tabular}
\end{table*}

Our training corpus pools a large collection of open-source and proprietary robot
demonstration datasets, listed in Table~\ref{tab:datasets_frames}. Our guiding
principle was to include all available data of reasonable quality and build a
pipeline that converts heterogeneous sources into a unified format. Below we
describe the conversion steps, the filtering criteria, and the final
representation choices.

Our in-house data consists of teleoperation demonstrations collected on the
AgiBot G2, AgiBot G2X, and PrimeBOT H2 humanoid platforms for household tasks
such as laundry and room tidying.

\section{Training Details}

\subsection{Architecture Hyperparameters}

\begin{table}[H]
  \centering
  \caption{Model architecture.}
  \label{tab:arch_hparams}
  \begin{tabular}{l l}
    \toprule
    Component & Value \\
    \midrule
    Backbone init & Wan2.2-TI2V-5B~\cite{wan} \\
    Parameters (total) & 5.05\,B \\
    Hidden dim & 3072 \\
    Transformer blocks & 30 \\
    Attention heads & 24 \\
    Patch size (spatial) & 2$\times$2 \\
    VAE spatial compression & 16$\times$ \\
    VAE temporal compression & 4$\times$ \\
    Latent channels & 48 \\
    Text encoder & umT5-XXL, frozen \\
    Action tokenizer & 2-layer MLP \\
    \bottomrule
  \end{tabular}
\end{table}

\subsection{Training Hyperparameters}
\label{sec:hyperparameters}

We report the architecture and the midtraining configuration for PrimeSim. All
downstream stages reuse the same optimizer, precision, and backbone unless noted
elsewhere.

The architecture is initialized from Wan2.2-TI2V-5B~\cite{wan} and retains its
backbone dimensions. We add only some lightweight action conditioning modules and three learnable view tokens on top of the frozen text encoder, so the trainable parameter count is close to the backbone's.

Midtraining uses a cosine-decayed learning rate with a short warmup and a small
minimum learning rate, which we found necessary to avoid late-stage loss spikes
on long training runs. The action classifier-free-guidance dropout is set to
0.1 to enable guidance at inference without retraining.

\begin{table}[H]
  \centering
  \caption{Midtraining hyperparameters. The architecture is initialized from the Wan2.2-TI2V-5B checkpoint, whose own training is not reported here.}
  \label{tab:pretrain_hparams}
  \begin{tabular}{l l}
    \toprule
    Hyperparameter & Value \\
    \midrule
    Optimizer & AdamW \\
    Learning rate & $1\times10^{-4}$ \\
    LR schedule & cosine decay \\
    Warmup steps & 10{,}000 \\
    Min LR ratio & 0.05 ($5\times10^{-6}$) \\
    $\beta_1, \beta_2$ & 0.9, 0.95 \\
    Weight decay & $1\times10^{-2}$ \\
    Precision & bfloat16 \\
    Action CFG dropout & 0.1 \\
    \midrule
    Resolution (per view) & 192$\times$256 \\
    Frames per sample & 33 \\
    Sampling FPS & 10 \\
    Train/val split & 0.99 / 0.01 \\
    \bottomrule
  \end{tabular}
\end{table}

\subsection{Distillation Hyperparameters}

Stage~3 refines the causal consistency student of Stage~2 with the asymmetric
DMD objective of Sec.~\ref{sec:dmd}. Three models take part. The \emph{teacher}
is the frozen bidirectional midtrained model and supplies the real-data score.
The \emph{student} is the causal few-step generator being distilled. The
\emph{critic} is initialised from the teacher and then trained with standard
flow matching on the student's own samples, so it tracks the distribution the
student currently produces rather than the one the teacher defines. Only LoRA
adapters are trained on the student, and the adapters are exported directly
rather than as an EMA of the full weights, which is why the adapter rank
appears in the table.

The critic is updated five times per generator update. It has to stay ahead of
a generator that is itself moving, and an under-trained critic supplies a
gradient that points at stale targets. The real guidance scale of 3.0 is
applied to the teacher's score, which sharpens the supervision the student is
matched against; without it the teacher averages over modes and the student
inherits that averaging. Both networks use the same learning rate, and the
sigma shift is the resolution-dependent noise-schedule shift of the flow
matching formulation. Because the surrogate loss is built so that its gradient
is exactly Eq.~\ref{eq:kl_grad}, the KL objective carries no separate
coefficient, and no balancing weight between the two losses appears in the
table. Table~\ref{tab:stage3_dmd_hparams} lists the settings used for every Stage~3
result reported in this work.

\begin{table}[H]
  \centering
  \caption{Stage~1 autoregressive modelling hyperparameters. The model is
  initialized from the midtrained PrimeSim checkpoint and adapted to
  chunk-causal generation using teacher forcing.}
  \label{tab:stage1_ar_hparams}
  \small
  \begin{tabular}{l l}
    \toprule
    Hyperparameter & Value \\
    \midrule
    Initialization & Midtrained PrimeSim \\
    Training objective & Flow matching with teacher forcing \\
    Trainable parameters & LoRA adapters \\
    Optimizer & AdamW \\
    Learning rate & $1\times10^{-4}$ \\
    LR schedule & Cosine decay \\
    Warmup steps & 10{,}000 \\
    Minimum LR ratio & 0.05 ($5\times10^{-6}$) \\
    $\beta_1,\beta_2$ & 0.9, 0.95 \\
    Weight decay & $1\times10^{-2}$ \\
    Precision & bfloat16 \\
    Action CFG dropout & 0.1 \\
    Resolution per view & $192\times256$ \\
    RGB frames per sample & 33 \\
    Latent frames per window, $T$ & 9 \\
    Sampling FPS & 10 \\
    Chunk size, $k$ & 3 latent timesteps \\
    Number of chunks, $T/k$ & 3 \\
    Within-chunk attention & Bidirectional \\
    Across-chunk attention & Causal \\
    Teacher sampling steps & 50 \\
    Train/validation split & 0.99 / 0.01 \\
    \bottomrule
  \end{tabular}
\end{table}

\begin{table}[H]
  \centering
  \caption{Stage~2 causal consistency distillation hyperparameters. Stage~2
  initializes both teacher and student from the Stage~1 causal model and
  distills the 50-step teacher into a four-step causal student using one online
  teacher ODE step per iteration.}
  \label{tab:stage2_cd_hparams}
  \small
  \begin{tabular}{l l}
    \toprule
    Hyperparameter & Value \\
    \midrule
    Initialization & Stage~1 causal model \\
    Training objective & Causal consistency distillation \\
    Trainable parameters & Student LoRA adapters \\
    Optimizer & AdamW \\
    Learning rate & $1\times10^{-4}$ \\
    Precision & bfloat16 \\
    Resolution per view & $192\times256$ \\
    RGB frames per sample & 33 \\
    Latent frames per window, $T$ & 9 \\
    Sampling FPS & 10 \\
    Chunk size, $k$ & 3 latent timesteps \\
    Number of chunks, $T/k$ & 3 \\
    Sampling steps & 4 \\
    \bottomrule
  \end{tabular}
\end{table}

\begin{table}[H]
  \centering
  \caption{Stage~3 asymmetric distribution matching distillation
  hyperparameters. Stage~3 initializes the generator from the four-step causal
  student produced by Stage~2 and refines it using a frozen bidirectional
  midtrained model as the real-score teacher and a trainable fake-score network
  that tracks the student's generated distribution.}
  \label{tab:stage3_dmd_hparams}
  \small
  \begin{tabular}{l l}
    \toprule
    Hyperparameter & Value \\
    \midrule
    Initialization & Stage~2 causal student \\
    Training objective & Asymmetric DMD \\
    Trainable parameters & Student LoRA adapters and fake-score network \\
    Optimizer & AdamW \\
    Learning rate & $1\times10^{-4}$ \\
    Precision & bfloat16 \\
    Resolution per view & $192\times256$ \\
    RGB frames per sample & 33 \\
    Latent frames per window, $T$ & 9 \\
    Sampling FPS & 10 \\
    Chunk size, $k$ & 3 latent timesteps \\
    Number of chunks, $T/k$ & 3 \\
    Sampling steps & 4 \\
    Critic\,:\,generator update ratio & $5:1$ \\
    Generator learning rate & $1\times10^{-4}$ \\
    Critic learning rate & $1\times10^{-4}$ \\
    Student LoRA rank & 32 \\
    \bottomrule
  \end{tabular}
\end{table}

\begin{table}[H]
  \centering
  \caption{Stage~4 episode-level rollout and hierarchical temporal memory
  hyperparameters. Cache capacities are measured in temporal latent slots.}
  \label{tab:stage4_memory_hparams}
  \small
  \begin{tabular}{l l}
    \toprule
    Hyperparameter & Value \\
    \midrule
    RGB frames per sample & 33 \\
    Latent frames per window, $T$ & 9 \\
    Sampling steps & 4 \\
    Global sink capacity & 3 \\
    Slow-memory capacity & 9 \\
    Local-memory capacity & 9 \\
    Total KV-cache capacity & 21 \\
    Generator updates per episode & 1 \\
    \bottomrule
  \end{tabular}
\end{table}
\section{Latency Benchmark Details}
\label{sec:latency_appendix}

This appendix collects the protocol details behind
Table~\ref{tab:inference_speed}: the timing environment, per-model warm-up and
chunk semantics, the configuration of every measured system, and the
latent-token proxy used for fairness discussion.

\subsection{Environment and Timing Scope}

\begin{table}[H]
  \centering
  \caption{Timing environment. All reported numbers use a single GPU at batch
  size one.}
  \label{tab:latency_env}
  \begin{tabular}{l l}
    \toprule
    Item & Value \\
    \midrule
    GPU & NVIDIA A100-SXM4-80GB \\
    Precision & bfloat16 \\
    Batch size & 1 (single episode / prompt stream) \\
    Timing API & \texttt{time.perf\_counter()} with \texttt{torch.cuda.synchronize()} \\
    Reported aggregate & steady-state mean, warm-up discarded \\
    \bottomrule
  \end{tabular}
\end{table}

The unit of work is one autoregressive chunk of future frames at each model's
native chunk size. The timed region covers the chunk generation call, including
GPU-to-CPU materialization of outputs when it is part of the public API:
\texttt{model.step()} including numpy conversion for GE-Sim 2.0,
\texttt{pipe()} plus \texttt{.cpu()} on future frames for BWM, text/observation
encoding, chain-of-thought sampling and VAE decoding with preloaded weights for
CausalWM, and \texttt{run\_episode} with checkpoint loading, text encoding and
forward kinematics excluded for PrimeSim. Checkpoint loading, tokenizer/model
construction and first-time CUDA kernel compilation are never timed; warm-up
chunks absorb compilation and cache effects.

\subsection{Warm-up, Repeats and Chunk Semantics}

\begin{table}[H]
  \centering
  \caption{Warm-up schedule, timed repetitions and aggregation per model.}
  \label{tab:latency_warmup}
  \setlength{\tabcolsep}{4pt}
  \begin{tabular}{l c c l}
    \toprule
    Model & Warm-up & Timed & Aggregation \\
    \midrule
    PrimeSim Stage-3 & 1 window & 3 repeats & mean \\
    GE-Sim 2.0 & 3 chunks & 12 chunks & mean of last 5 (steady state) \\
    BWM (fair-aligned) & 1 chunk & 3 chunks & mean \\
    BWM (official) & 1 chunk & 2 chunks & mean \\
    Ctrl-World & bench default & 5 chunks & mean \\
    CausalWM & weights preloaded & 2 chunks & mean \\
    \bottomrule
  \end{tabular}
\end{table}

\begin{table}[H]
  \centering
  \caption{Chunk semantics per model: conditioning history, future frames per
  chunk $T$, number of camera views $V$ and denoising steps.}
  \label{tab:latency_chunks}
  \setlength{\tabcolsep}{4pt}
  \begin{tabular}{l l c c l}
    \toprule
    Model & History & Future $T$ & Views $V$ & Denoise steps \\
    \midrule
    PrimeSim Stage-3 & continuous KV & 32 (33-frame window) & 3 & 4 (1000, 750, 500, 250) \\
    GE-Sim 2.0 & 4 (memory band) & 25 & 3 & 4 (distilled) \\
    BWM (fair-aligned) & 4 & 25 & 1 & 50 \\
    BWM (official) & 9 & 48 & 1 & 50 \\
    Ctrl-World & 6 & 5 & 3 (stacked) & 50 \\
    CausalWM & first frame + text & 25 & 1 & 4+4+4 per stage \\
    \bottomrule
  \end{tabular}
\end{table}

\subsection{Per-model Configurations}

\paragraph{PrimeSim Stage-3 (ours).}
Wan2.2-TI2V-5B with Stage-3 asymmetric DMD, action-conditioned, continuous KV
cache with local attention span 9. Three views at $288\times384$, 33-frame
windows (32 usable frames) under the RoboTwin~2.0 Stage-3 evaluation contract,
four-step consistency sampler at sigmas $(1000, 750, 500, 250)$. Steady-state
repeats measure 10.006, 10.011 and 10.329 TempFPS (mean 10.115, 3.16\,s per
window); a longer continuous-KV episode of about 72 model frames measures 7.63
TempFPS, and a cold-start run without warm-up measures 6.41 TempFPS --- neither
is used as the headline. Timed scope is \texttt{run\_episode} with loading,
text encoding and forward kinematics excluded.

\paragraph{GE-Sim 2.0.}
Cosmos 2B with Pose Expert, released distilled checkpoint, action-conditioned
with an episode band of intrinsics, extrinsics and trajectory. Three views at
$384\times512$, history 4, future 25, four-step distilled sampling with sigma
schedule $(1.0, 0.9375, 0.8333, 0.625)$, \texttt{triton\_rope} enabled and
Liger/SpargeAttention off. Twelve timed chunks follow three warm-up chunks;
early chunks are noisy (9.78\,s to 44.84\,s), so the steady state is the mean
of the last five, 11.15\,s per chunk.

\paragraph{BWM, fair-aligned.}
Wan2.2-TI2V-5B with action AdaLN (BLM checkpoint, step-12000), conditioned on
absolute end-effector actions. Single view at $384\times512$, history 4, future
25 (chunk of 29 frames total), 50-step sampling at CFG 1.0. Three timed chunks
measure 6.23, 6.15 and 6.19\,s (mean 6.19\,s). The distilled BWM-fast weights
were not publicly available at measurement time.

\paragraph{BWM, official configuration.}
Vendor-default inference config: single view at $672\times896$, history 9,
future 48, 50-step sampling. Mean chunk time 61.71\,s. Reported as the heavy
vendor-default point rather than as a fair peer.

\paragraph{Ctrl-World.}
SVD image-to-video UNet with CLIP text, conditioned on 7D actions. Three views
stacked on the height axis into a $576\times320$ canvas ($192\times320$ per
view), history 6, future 5, 50-step sampling. Five timed chunks measure 5.76 to
5.80\,s (mean 5.78\,s).

\paragraph{CausalWM.}
LTX-2.3 22B with Gemma-3-12B, generating a flow $\to$ pointmap $\to$ RGB
chain-of-thought from text and a first frame --- not action-conditioned. Single
view at $480\times640$, 25 future frames, four steps per chain-of-thought
stage. Two timed chunks measure 39.23 and 36.72\,s (mean 37.97\,s), with
weights preloaded and load time excluded.

\subsection{Latent-token Proxy}

For fairness discussion we estimate the number of latent tokens per forward as
\begin{equation}
\widehat{\#\text{tokens}} \;\approx\; V \times
\Big(\big\lfloor\tfrac{T_{\text{lat}}-1}{4}\big\rfloor + 1\Big) \times
\big\lfloor\tfrac{H}{8}\big\rfloor \times \big\lfloor\tfrac{W}{8}\big\rfloor ,
\label{eq:token_proxy}
\end{equation}
i.e.\ a spatial$\times$temporal compression grid in the style of Cosmos/Wan
tokenizers (spatial stride 8, temporal compression $\approx 4$). This is a
coarse grid proxy, not FLOPs.

\begin{table}[H]
  \centering
  \caption{Approximate latent tokens per forward (Eq.~\ref{eq:token_proxy}).
  GE-Sim 2.0 counts its effective temporal extent including memory; Ctrl-World
  uses its stacked $576\times320$ canvas; CausalWM's temporal stride may
  differ, so its value is an estimate.}
  \label{tab:token_proxy}
  \begin{tabular}{l r}
    \toprule
    Model & $\widehat{\#\text{tokens}}$ \\
    \midrule
    PrimeSim Stage-3 & 46{,}656 \\
    GE-Sim 2.0 & 73{,}728 \\
    BWM (fair-aligned) & 24{,}576 \\
    BWM (official) & 141{,}120 \\
    Ctrl-World & 14{,}400 \\
    CausalWM & $\sim$19{,}200 \\
    \bottomrule
  \end{tabular}
\end{table}

PrimeSim's proxy uses the 33-frame window
($3 \times 9 \times 36 \times 48 = 46{,}656$); the continuous KV cache bounds
the attention span but does not change this per-window grid count, so it is not
inflated to a memory-plus-future figure.

\end{document}